\documentclass[letterpaper]{article} 
\usepackage{aaai2026}  
\usepackage{times}  
\usepackage{helvet}  
\usepackage{courier}  
\usepackage[hyphens]{url}  
\usepackage{graphicx} 
\usepackage{natbib}  
\usepackage{caption} 
\usepackage{algorithm}
\usepackage{algorithmic}
\usepackage{amsmath}
\usepackage{booktabs}
\usepackage{multirow}
\usepackage{subcaption}
\usepackage{latexsym}
\usepackage{algorithm}
\usepackage{algorithmic}
\usepackage{amsmath}
\usepackage{booktabs}
\usepackage{multirow}
\usepackage{subcaption}
\usepackage{newfloat}
\usepackage{listings}
\usepackage{tcolorbox}
\usepackage{array}       
\usepackage{adjustbox}
\usepackage{float}
\usepackage{newfloat}
\usepackage{listings}
\DeclareCaptionStyle{ruled}{labelfont=normalfont,labelsep=colon,strut=off} 
\floatstyle{ruled}
\newfloat{listing}{tb}{lst}{}
\floatname{listing}{Listing}
\title{Latent Self-Consistency for Reliable Majority‑Set Selection in Short‑ and Long‑Answer Reasoning}

\author {
    Jungsuk Oh,
    Jay-Yoon Lee\thanks{Corresponding author: lee.jayyoon@snu.ac.kr}
}
\affiliations {
    Graduate School of Data Science, Seoul National University\\
    \{luke0112, lee.jayyoon\}@snu.ac.kr
}

\begin{document}

\maketitle

\begin{abstract}
Probabilistic decoding in Large Language Models (LLMs) often yields inconsistent outputs, particularly on complex or long-form questions. Self-Consistency (SC) mitigates this for short-form QA by majority voting over exact strings, whereas Universal Self-Consistency (USC) and Weighted Unigram Consistency Score (WUCS) extend to long-form responses but lose accuracy on short-form benchmarks. 

We introduce \textbf{Latent Self-Consistency (LSC)}, which selects the most semantically consistent response using learnable token embeddings.
LSC's lightweight forward processing of summary tokens only introduces negligible runtime overhead (at most $0.9\%$) on top of standard decoding of 
the base LLM, and requires no changes to the model architecture.

Across 6 short-form and 5 long-form reasoning benchmarks (e.g., MATH, MMLU, TruthfulQA), LSC surpasses SC, USC, and WUCS on both short-form and long-form on average performance, while adding negligible computational overhead on vanilla inference. These results position LSC as a reliable consistency-selection method that works effectively across various answer formats.
Additionally, LSC provides well-calibrated confidence estimates, maintaining low expected calibration error across both answer formats.
\end{abstract}

\begin{links}
    \link{Code}{https://github.com/jeongseokO/LatentSC}
    \link{Datasets}{https://huggingface.co/jeongseokoh}
\end{links}

\section{Introduction}

Large Language Models (LLMs) have achieved remarkable success across diverse natural language processing tasks, from mathematical reasoning to code generation \cite{brown-etal-2020-language, chowdhery-etal-2022-palm, openai2023gpt4}. However, as inherently probabilistic generators, they frequently yield divergent outputs on repeated runs, undermining consistency and reliability in downstream applications. To mitigate this variability, Self-Consistency (SC) \cite{wang-etal-2022-self} generates multiple candidate responses through independent sampling and applies majority voting to select the most frequent answer. This mechanism has proven highly effective, achieving substantial accuracy improvements in mathematical reasoning and multiple-choice benchmarks by filtering out anomalous reasoning paths.

SC's reliance on exact string matching, however, severely limits its applicability. While effective for ``short-answer'' tasks such as numerical problems or multiple-choice questions where alternative surface forms are rare, SC fails in ``long-answer'' scenarios including code generation, summarization, or detailed explanations, where semantically equivalent responses often differ lexically and structurally. To address this limitation, several consistency-based extensions have emerged. Universal Self-Consistency (USC) \cite{chen-etal-2023-universal} employs the LLM itself as a semantic judge by concatenating all candidates into a single prompt for evaluation; however, this approach incurs substantial computational overhead (approximately 10\% additional inference time) and significant memory usage (around 15\% additional memory), while introducing potential judge-model biases. Weighted Unigram Consistency Score (WUCS) \cite{jain-etal-2024-lightweight} offers a lightweight alternative using term-frequency–weighted Jaccard similarity with probabilistic weighting, but sacrifices accuracy on short-answer tasks and fails to capture semantic coherence in complex reasoning. {These alternative methods sometimes achieve competitive accuracy, but such gains only partly arise from putting together truly consistent reasoning paths. For example, USC often relies on LLM reasoning when it selects a more accurate response that disagrees with the majority response.
}

{In practice, this leads to an unsatisfying trade-off: SC performs strongly on short-answer benchmarks but breaks down on long-form tasks, whereas USC and WUCS are better suited to free-form generation yet often degrade performance on short-answer evaluation. Empirically, our experiments (Appendix~K) show that, given only the input question, it is difficult to reliably predict whether the model will produce a short-answer–style or a long-answer–style response, making it impractical to route queries to different decoding schemes based on an assumed answer format.}

\begin{figure*}[!htb]
  \centering
  \includegraphics[width=0.95\textwidth]{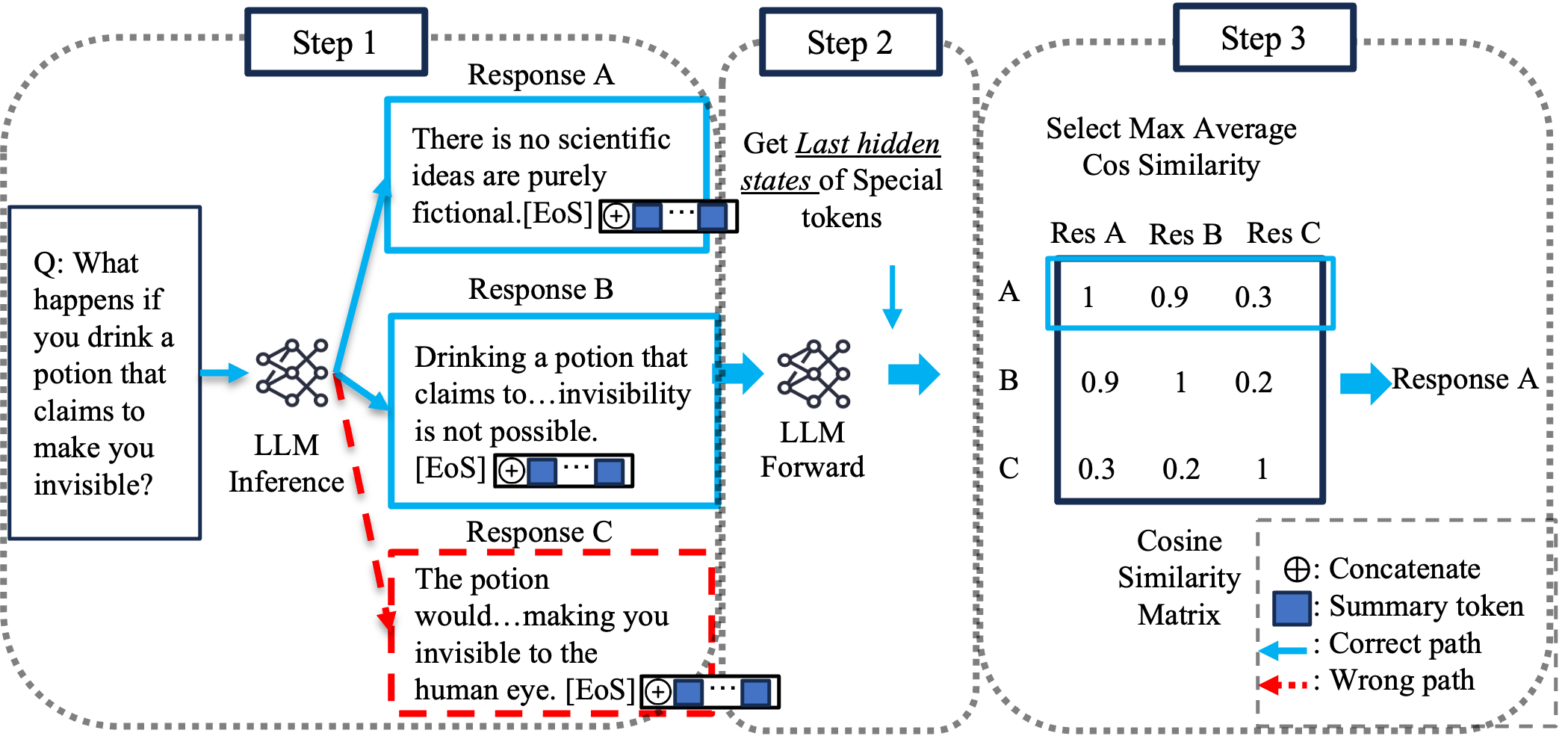}
  \caption{Overview of the Latent Self-Consistency (LSC) inference framework.}

  \label{fig:overview}
\end{figure*}
In this work, we introduce \emph{Latent Self-Consistency} (LSC), a framework that aims to achieve the best of both worlds: matching SC's high accuracy on short-answer tasks while extending robustly to long-answer scenarios, with minimal additional computational cost. LSC appends a small set of trainable summary tokens after the EOS token of each generated response and reuses the key–value (KV) cache from the original generation to compute their final-layer representations, yielding compact embeddings that capture the semantic essence of each response.{
{LSC introduces a negligible runtime overhead (at most 0.9\%)
compared to standard decoding }
and requires no changes to the model architecture.} By computing pairwise cosine similarities with these learned embeddings,
LSC identifies the most consistent response without constructing new prompts or performing any complex text processing.

Our contributions are {as follows}:
\begin{itemize}
    \item \textbf{Universal majority selection across answer formats.} 
    {We present the first consistency-based selection method 
    that delivers strong performance on both short- and long-answer tasks
    while using a single unified approach, with no format-dependent tuning or heuristics.}
    
    \item \textbf{Efficient semantic consistency via learned suffix embeddings.} By encoding response semantics into compact learned embeddings obtained from a few summary tokens, LSC attains USC-level semantic discrimination with substantially reduced {runtime} and memory overhead, making it practical for real-time deployment.

    \item \textbf{Reliable and calibrated confidence estimates across tasks.} LSC enables 
    uncertainty quantification, achieving {competitive expected calibration error (ECE) scores for both short- and long-answer benchmarks.}
    
    \item \textbf{Dynamic boundary detection for robust majority set identification.} We introduce a dynamic Top-$K$ boundary detection algorithm that automatically filters out noisy outliers, {further enhancing selection accuracy
    by focusing on the most semantically coherent subset.
    }
\end{itemize}

\section{Related Works}

\subsection{Consistency‐Based Selection in LLMs}
Self‐Consistency (SC) \cite{wang-etal-2022-self} generates $N$ independent samples and selects the most frequent answer via exact string matching, yielding strong gains on "short answer form" tasks but failing when semantically equivalent outputs vary lexically.  
Universal Self‐Consistency (USC) \cite{chen-etal-2023-universal} uses the LLM itself as a referee by concatenating all $N$ candidates into a single prompt and performing one additional forward pass. However, this approach requires constructing entirely new prompts, preventing KV cache reuse and introducing substantial latency and potential judge-model bias.  
Unigram Consistency Score (UCS) and its weighted variant WUCS \cite{jain-etal-2024-lightweight} compute probability-weighted overlap measures for text similarity assessment, which while efficient, struggle to capture deep chain-of-thought coherence.
Fine‐grained Self‐Consistency (FSC) \cite{wang-etal-2024-integrate} integrates segment-level consensus by re-generating and synthesizing responses from scratch, incurring extra latency due to these additional generative passes.  
In contrast, our LSC extracts small, trainable token embeddings by simply appending learnable tokens after the EOS token of each generated response, enabling KV cache reuse for efficient forward passes. This approach unifies SC's accuracy on short-answer tasks with robust, low-latency selection for long-answer tasks.
\begin{table*}[!htb]
\centering
\small
\begin{tabular}{ll cc cc cc}
\toprule
& & \multicolumn{1}{c}{\textbf{Short}} & \multicolumn{1}{c}{\textbf{Long}} & \multicolumn{2}{c}{\textbf{Overhead on top of Vanilla}} \\
\cmidrule(lr){3-3} \cmidrule(lr){4-4} \cmidrule(lr){5-6}
\textbf{Models} & \textbf{Methods} 
& \textbf{Average(\%)} 
& \textbf{Average(\%)} 
& \textbf{Time(\%)} 
& \textbf{Memory(\%)} 
& \\
\midrule
\multirow{5}{*}{\shortstack[l]{LLaMA3.1‑\\8B-Instruct}} 
& SC                            
  & 72.2     & N/A     & 2E-3    & 0.0  \\
& WUCS                        
  & 70.0      & 63.8       & 3E-2    & 0.0   \\
& USC                             
  & 67.7     & 64.8       & 7.4     & 16.2  \\
& LSC(Ours)                     
  & \underline{72.3}  & \underline{65.2}    & {0.2}  & {5E-3}  \\
&  \quad +Dynamic TopK                          
  & \textbf{72.8}     & \textbf{65.8}       & {0.2}     & {5E-3} \\
\midrule
\multirow{5}{*}{Qwen3-8B} 
& SC                            
  & \textbf{77.1}     & N/A     & 1E-3    & 0.0   \\
& WUCS                        
  & {74.9}      & \textbf{75.4}       & 2E-2  & 0.0  \\
& USC                              
  & 75.4     & 74.7       & 18.2     & 20.2   \\
& LSC(Ours)                         
  & \underline{76.8}  & \underline{74.9}    & 0.6  & 5E-3  \\
&  \quad +Dynamic TopK                          
  & {76.5}     & \underline{74.9}     & 0.6     & 5E-3 \\
\midrule

\multirow{5}{*}{\shortstack[l]{LLaMA3.3‑\\70B-Instruct}} 
& SC                            
  & \textbf{82.7}     & N/A     & 1E-4    & 0.0   \\
& WUCS                        
  & 81.1      & 76.7       & 8E-3    & 0.0   \\
& USC                             
  & {82.5}     & {76.8}       &  9.7    & 7.5  \\
& LSC(Ours)                    
  & 82.6  & \textbf{77.6}    & 0.9  & {1E-3}  \\
&  \quad +Dynamic TopK                          
  & \textbf{82.7}     & \textbf{77.6}     & 0.9     & {1E-3}  \\
\bottomrule
\end{tabular}%

\caption{
Overall performance–efficiency trade-off of LSC. Time/Memory columns report percentage overhead over vanilla inference when generating 10 candidates.
}

\label{tab:overall-table}
\end{table*}

\subsection{Prompt Tuning and Learnable Token Embeddings}
Existing prompt tuning approaches \cite{lester-etal-2021-power, li-etal-2021-prefix} typically append learnable tokens to user prompts for domain adaptation without full model fine-tuning. These methods operate in a \textbf{pre-generation} manner, modifying the input before the model generates responses. 

Our work introduces the first approach to learn semantic summary representations of generated responses through learnable token embeddings. Our LSC method operates in a \textbf{post-generation} manner, appending learnable tokens after the assistant's response generation is complete to capture the semantic essence of each response. We refer to these as "\emph{summary tokens}" since they are trained to encode compact semantic summaries of the full responses. Crucially, since we only update the summary token embeddings while keeping the base LLM frozen, our method preserves the model's original natural language capabilities without any degradation. This design enables efficient semantic consistency evaluation through lightweight forward passes while maintaining complete separation between consistency measurement and the model's core competencies.


\section{Method}
\begin{table*}[!htb]
\small
\setlength{\tabcolsep}{4pt}
\centering
\begin{tabular}{llccccccc}
\toprule
& & \multicolumn{4}{c}{\textbf{In-domain}} & \multicolumn{2}{c}{\textbf{Out-of-domain}} & \multirow{2}{*}{\textbf{Average}} \\
\cmidrule(lr){3-6} \cmidrule(lr){7-8}
\textbf{Models} & \textbf{Decoding Methods} 
& \textbf{GSM8K} 
& \textbf{MATH} 
& \textbf{TriviaQA} 
& \textbf{MMLU} 
& \textbf{TruthfulQA MC1} 
& \textbf{CommonsenseQA} \\
\midrule
\multirow{6}{*}{\shortstack[l]{LLaMA3.1\\8B-Inst.}}
& Vanilla (1 path)            & 83.2 & 39.4 & 62.4 & 68.8 & 57.2 & 74.9 & 64.3 \\
& Self Consistency            & \underline{92.2} & 52.5 & 72.7 & 73.6 & \textbf{63.0} & \textbf{79.4} & 72.2 \\
& WUCS                        & 89.8 & 45.6 & 73.2 & 72.8 & 61.8 & 76.6 & 70.0 \\
& USC                         & 86.2 & 44.8 & 70.1 & 71.1 & 59.6 & 74.6 & 67.7 \\
& LSC (Ours)                  & \textbf{92.3} & \underline{52.6} & \underline{74.1} & \underline{73.8} & 62.1 & 79.0 & \underline{72.3} \\
& \quad +Dynamic TopK         & \underline{92.2} & \textbf{52.9} & \textbf{75.7} & \textbf{73.9} & \underline{62.7} & \underline{79.1} & \textbf{72.8} \\
\midrule
\multirow{6}{*}{Qwen3-8B}
& Vanilla (1 path)            & 92.8 & 71.9 & 48.1 & 76.7 & 72.6 & 81.6 & 74.0 \\
& Self Consistency            & \textbf{94.5} & \textbf{76.7} & \textbf{56.1} & \textbf{77.2} & \underline{75.2} & \underline{82.6} & \textbf{77.1} \\
& WUCS                        & 93.8 & 72.1 & 49.9 & 76.8 & 75.0 & 81.9 & 74.9 \\
& USC                         & 93.7 & 75.6 & 50.9 & 76.0 & 74.4 & 81.8 & 75.4 \\
& LSC (Ours)                  & \textbf{94.5} & 75.9 & \underline{55.5} & \underline{77.1} & 74.9 & \textbf{82.7} & \underline{76.8} \\
& \quad +Dynamic TopK         & 94.4 & \underline{76.6} & 53.2 & 77.0 & \textbf{75.4} & \underline{82.6} & 76.5 \\
\midrule
\multirow{7}{*}{\shortstack[l]{LLaMA3.3\\70B-Inst.}}
& Vanilla (1 path)            & 95.9 & 65.1 & 85.2 & 84.1 & 73.9 & 84.4 & 81.4 \\
& Self Consistency            & \underline{96.9} & 68.6 & 86.3 & \textbf{84.6} & 75.0 & \underline{84.7} & \textbf{82.7} \\
& WUCS                        & 95.5 & 63.7 & 84.6 & 84.3 & 74.2 & 84.1 & 81.1 \\
& USC                         & 95.8 & 68.2 & \textbf{86.5} & \textbf{84.6} & 74.8 & \textbf{84.9} & 82.5 \\
& LSC (Ours)                  & \underline{96.9} & \textbf{68.7} & 86.3 & 84.2 & \textbf{75.3} & 84.3 & 82.6 \\
& \quad +Dynamic TopK         & \textbf{97.0} & \textbf{68.7} & \underline{86.4} & 84.2 & \textbf{75.3} & 84.5 & \textbf{82.7} \\
\bottomrule
\end{tabular}
\caption{{Performance comparisons on short-answer benchmarks.
Only \textbf{In-domain} sets were used for summary-token training. }
The highest score in each column is in \textbf{bold}, and the second-highest is \underline{underlined}. All scores are reported as percentages.}
\label{tab:short-answer-table}
\end{table*}
\subsection{Inference Phase}

Figure~\ref{fig:overview} illustrates the Latent Self-Consistency (LSC) inference pipeline in three phases: Candidate generation, Post-generation encoding, and Majority answer detection. 

For (1) Candidate generation, given an input question we prompt the base LLM and sample $N$ candidate responses. 
For (2) Post-generation encoding, we append $K$ learnable summary tokens at the end of each response and run a lightweight forward pass to produce {response} representations $\{z_i\}$. Unlike USC, which concatenates all responses into a new prompt and recomputes the entire sequence, LSC appends only a few tokens at the end of each candidate and reuses the existing key–value (KV) cache from the candidate generation phase. As a result, the model needs to compute representations only for the $K$ additional tokens rather than processing entire sequences from scratch. {This makes LSC highly efficient as it only adds negligible overhead on top of base LM: at most $0.9\%$ runtime and $0.005$\% memory overhead in our experiments.} 

For (3) Majority answer detection, we first compute the cosine similarity matrix between all pairs of response embeddings, 
$S_{ij} = \frac{z_i\cdot z_j}{\|z_i\|\|z_j\|}$. {We then assign an exponentially weighted similarity score to each response $i$,}
\[
w_i = \frac{1}{N-1}\sum_{j\neq i}\exp\bigl(S_{ij}/\tau'\bigr),
\]
with $\tau' = 0.5$. A simple arithmetic mean of $\{S_{ij}\}_{j\neq i}$ treats all the responses equally 
{and will likely pull} $w_i$ toward the global centroid, especially in the presence of outliers. 
{In contrast, the exponential weighting effectively suppresses low-similarity responses, resulting in $w_i$ being primarily determined by candidates from the majority set.}

To further suppress the influence of residual outliers, we additionally propose a Dynamic Top-$K$ scheme that estimates the size of the majority set directly from the similarity structure. For each $K \geq 2$, let $\sigma_i$ sort other responses in decreasing similarity to $i$, and define the average similarity to the $K$ nearest neighbors
\[
w_i^{(K)} = \frac{1}{K}\sum_{j=1}^K S_{i,\sigma_i(j)},
\]
with the best score at that neighborhood size
\[
w^{(K)}_{\max} = \max_i w_i^{(K)}.
\]
We then track the drop between $K$ and $K-1$,
\[
\Delta^{(K)} = w^{(K-1)}_{\max} - w^{(K)}_{\max},
\]
and choose
\[
K^* = \arg\max_K \Delta^{(K)} - 1,
\]
which corresponds to the neighborhood size just before the largest decrease in maximum similarity. Finally, we select the index
\[
i^* = \arg\max_i w_i^{(K^*)}
\]
as the majority choice.

\subsection{Training Phase}
\subsubsection{Dataset Construction}
We assemble a balanced mix of short‐answer and long‐answer samples using LLM‐based pseudo‐labeling. For each question, an LLM generates an answer from which we extract the answer string as the label, identifying which responses converge on the same conclusion. Short‐answer data are drawn from GSM8K \cite{cobbe2021training}, MATH \cite{hendrycks2021measuring}, and TriviaQA \cite{joshi2017triviaqa}. Long‐answer samples are derived from MMLU \cite{hendrycks2020measuring} by prompting chain‐of‐thought rationales and removing the short-form answer to yield 'free‐form explanations with known pseudo-labels'. We maintain a 50:50 ratio of short to long formats. Detailed dataset construction procedures, statistics, and examples are provided in Appendix A, with prompt templates in Appendix B.

\subsubsection{Learning Representations via Supervised Contrastive Learning}
To bring semantically consistent responses closer together, we use supervised contrastive learning: embeddings of responses sharing the same answer label are pulled together, while those with different labels are pushed apart.  Concretely, we append \(K\) learnable special tokens after each response and fine‐tune only their embeddings (e.g.\ \(K=6\), \(d=4096\) for 8B LLM, updating \(<3 \times 10^{-6}\) of parameters).  Let \(\{h_{i,1},\dots,h_{i,K}\}\) be the final‐layer representations of those tokens for response \(i\); we compute
\[
\bar h_i = \frac{1}{K}\sum_{m=1}^K h_{i,m}, 
\quad
z_i = \frac{\bar h_i}{\|\bar h_i\|}.
\]
With \(y_i\) the answer label, the supervised contrastive loss~\cite{he2020momentum,khosla2020supervised} over a batch of size \(N\) is
\[
\mathcal{L}_{\mathrm{SCL}}
= -\frac{1}{N}\sum_{i=1}^N \frac{1}{|\mathcal{I}_i^+|}\sum_{j\in\mathcal{I}_i^+}
\log\frac{\exp(z_i\!\cdot\!z_j/\tau)}
{\sum_{k\neq i}\exp(z_i\!\cdot\!z_k/\tau)},
\]
where \(\mathcal{I}_i^+ = \{j\neq i: y_j = y_i\}\) and \(\tau=0.07\).

\begin{figure}[t]
\centering
\includegraphics[width=0.8\columnwidth]{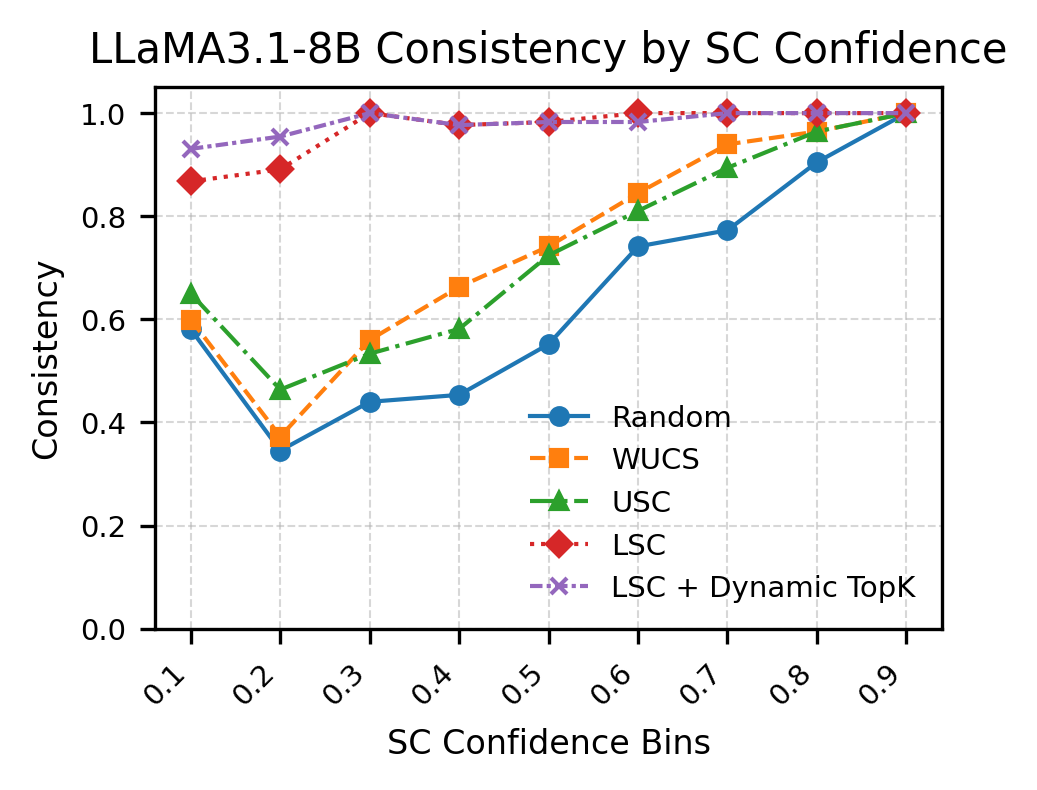}
\caption{
Consistency (fraction of examples where the selected response is in the true majority set) vs. majority-set size on MATH with LLaMA3.1-8B-Instruct, showing LSC outperforming all baselines and Dynamic Top-$K$ giving additional gains for small majority sets.
}
\label{majority-conf-graph}
\end{figure}

\begin{table*}[!htb]
\centering
\begin{subtable}{\textwidth}
\centering
\small
\begin{tabular}{ll cc cc cc cc}
\toprule
\multirow{2}{*}{\textbf{Models}} 
& \multirow{2}{*}{\textbf{Decoding Methods}} 
& \multicolumn{6}{c}{\textbf{MSMARCO v2.1 Natural Language Generation}} 
& \multicolumn{1}{c}{\textbf{TruthfulQA}} 
\\
\cmidrule(lr){3-8} \cmidrule(lr){9-9}
& 
& \textbf{Semantic Similarity}
& \textbf{BLEU 1} 
& \textbf{BLEU 2} 
& \textbf{BLEU 3} 
& \textbf{BLEU 4} 
& \textbf{ROUGE L}
& \textbf{Truth(\%)} \\
\midrule
\multirow{5}{*}{\shortstack[l]{LLaMA3.1‑\\8B-Instruct}} 
& Vanilla (1 path)                            
  & 0.8805   & {0.4122}   & 0.3137   & 0.2488    & 0.2005   & {0.5075} & 39.9\\
& WUCS                        
  & {0.8815}  & 0.3904   & 0.2996   & 0.2392 & 0.1938  & 0.4987 & 42.8 \\
& USC                              
  & {0.8812}     & 0.4065    & {0.3103}   & {0.2463}  & {0.1989}  & {0.5086} & \underline{44.2}\\
& LSC(Ours) 
    & \underline{0.8838}  & \underline{0.4269}   & \underline{0.3274}    & \underline{0.2613} & \underline{0.2118}  & \underline{0.5207} & 43.5 \\
&  \quad +Dynamic TopK    
    & \textbf{0.8861}  & \textbf{0.4336}   & \textbf{0.3333}    & \textbf{0.2667} & \textbf{0.2167}  & \textbf{0.5270} & \textbf{44.4}\\
\midrule
\multirow{5}{*}{Qwen3-8B} 
& Vanilla (1 path)                            
  & \textbf{0.8842}     & {0.3750}     & {0.2876}    & {0.2303}    & {0.1875}  & {0.4947} & 43.6 \\
& WUCS                        
  & 0.8835  & 0.3620   & 0.2788    & 0.2242 & 0.1835  & 0.4886 & \underline{44.7} \\
& USC                              
  & {0.8824}     & 0.3608    & 0.2772   & 0.2223  & 0.1812  & 0.4871 & 43.1 \\
& LSC(Ours) 
    & \underline{0.8838}  & \underline{0.3756}   & \underline{0.2896}    & \textbf{0.2330} & \textbf{0.1906}  & \textbf{0.4964} & \textbf{44.9} \\
&  \quad +Dynamic TopK    
    & {0.8836}  & \textbf{0.3761}   & \textbf{0.2897}    & \textbf{0.2330} & \underline{0.1905}  & \underline{0.4958} & 44.6 \\
\midrule
\multirow{5}{*}{\shortstack[l]{LLaMA3.3‑\\70B-Instruct}}  
& Vanilla (1 path)                            
  & \textbf{0.8926}     & {0.4694}     & {0.3662}    & {0.2981}    & {0.2463}  & {0.5618} & 48.0 \\
& WUCS                        
  & 0.8914  & 0.4277   & 0.3343    & 0.2716 & 0.2242  & 0.5500 & 48.7 \\
& USC                              
  & {0.8913}     & 0.4623    & 0.3603   & 0.2930  & 0.2423  & 0.5596 & \textbf{50.6} \\
& LSC(Ours) 
    & {0.8918}  & \underline{0.4724}   & \underline{0.3698}    & \underline{0.3016} & \underline{0.2499}  & \underline{0.5629} & 48.8 \\
&  \quad +Dynamic TopK    
    & \underline{0.8922}  & \textbf{0.4768}   & \textbf{0.3733}    & \textbf{0.3042} & \textbf{0.2519}  & \textbf{0.5652} & \underline{49.3} \\
\bottomrule
\end{tabular}%

\caption{Results on free-form generation QA tasks MSMARCO-NLG and TruthfulQA.}
\label{tab:long-qa-table}
\end{subtable}

\vspace{0.3cm}

\begin{subtable}{\textwidth}
\centering
\small
\setlength{\tabcolsep}{3.5pt}
\begin{tabular}{ll cc cc cc cc cc cc}
\toprule
\multirow{2}{*}{\textbf{Models}} 
& \multirow{2}{*}{\textbf{Decoding Methods}} 
& \multicolumn{1}{c}{\textbf{HumanEval}} 
& \multicolumn{1}{c}{\textbf{HumanEval+}} 
& \multicolumn{1}{c}{\textbf{MBPP}} 
& \multicolumn{1}{c}{\textbf{MBPP+}}
& \multicolumn{4}{c}{\textbf{CNN Dailymail (Summarization Task)}} \\
\cmidrule(lr){3-4}\cmidrule(lr){5-6}\cmidrule(lr){7-10}
& 
& \textbf{Pass@1}
& \textbf{Pass@1}   
& \textbf{Pass@1}
& \textbf{Pass@1}
& \textbf{ROUGE 1} 
& \textbf{ROUGE 2}
& \textbf{ROUGE L}
& \textbf{BertScore}\\
\midrule
\multirow{5}{*}{\shortstack[l]{LLaMA3.1‑\\8B-Instruct}} 
& Vanilla (1 path)                            
  & 54.47       & 47.13    & 64.83  & 54.50   & {0.3091}   & 0.1087   & {0.2077}   & {0.8699}   \\
& WUCS                        
  & {55.50}  & 51.00 & {67.03}   & 55.10  & 0.3033  & {0.1113}   & 0.2033   & 0.8689  \\
& USC                            
  & {56.93}     & 50.80     & {68.43} & {58.57} & 0.3033     & 0.1048    & 0.2013   & 0.8689  \\
& LSC(Ours)                         
  & \underline{57.93}   & \underline{51.23}  & \underline{69.23}   & \underline{58.90}  & \textbf{0.3173}  & \textbf{0.1127}   & \underline{0.2130}    & \underline{0.8712} \\
&  \quad +Dynamic TopK    
  & \textbf{58.50}   & \textbf{51.80}  & \textbf{69.30}   & \textbf{60.60}  & \underline{0.3168}  & \underline{0.1124}   & \textbf{0.2149}    & \textbf{0.8713} \\
\midrule
\multirow{5}{*}{Qwen3-8B} 
& Vanilla (1 path)                            
  & {77.80}     & 73.40    & 81.03   & 69.23   & {0.2978} & {0.0977} & {0.1968} & {0.8683}  \\
& WUCS                        
  & \textbf{80.30}   & \textbf{75.20}   & \underline{82.43}   & \underline{70.17}    & 0.2904  & 0.0966   & 0.1897    & 0.8668 \\
& USC                            
  & \underline{78.03}       & {73.17}       & \textbf{82.97}   & \textbf{71.00}    & 0.2864     & 0.0950    & 0.1888   & 0.8660 \\
& LSC(Ours)                         
  & \underline{79.30}   & \underline{74.20}  & {81.20}    & {69.40}   & \underline{0.3033}  & \textbf{0.1007}  & \textbf{0.1997}  & \textbf{0.8691} \\
&  \quad +Dynamic TopK    
  & {78.57}   & {73.50}  & {82.00}   & {70.10}  & \textbf{0.3034}  & \underline{0.1000}   & \underline{0.1992}    & \underline{0.8690} \\
\midrule
\multirow{5}{*}{\shortstack[l]{LLaMA3.3‑\\70B-Instruct}} 
& Vanilla (1 path)                            
  & \underline{82.30}       & \textbf{76.80}    & 84.90  & 68.31   & {0.2536}   & 0.0930   & {0.1700}   & {0.8609}   \\
& WUCS                        
  & {81.67}  & {73.21} & \underline{87.33}   & 70.60  & 0.2418  & {0.0902}   & 0.1626   & 0.8595  \\
& USC                            
  & {79.93}     & {74.44}     & {86.20} & {71.42} & 0.2538     & \underline{0.0941}    & 0.1699   & 0.8610  \\
& LSC(Ours)                         
  & \underline{82.30}   & {75.00}  & \textbf{87.80}   & \textbf{73.81}  & \underline{0.2540}  & {0.0939}   & \underline{0.1703}    & \underline{0.8611} \\
&  \quad +Dynamic TopK    
  & \textbf{83.51}   & \underline{75.57}  & {86.82}   & \underline{72.50}  & \textbf{0.2558}  & \textbf{0.0947}   & \textbf{0.1714}    & \textbf{0.8613} \\
\bottomrule
\end{tabular}%
\caption{Performance on code generation (HumanEval+, MBPP+) and summarization (CNN/DailyMail) tasks.}
\label{tab:coding-summary-comparison}
\end{subtable}

\caption{Long-answer task performance. The best result in each column is \textbf{bold} and the second-best is \underline{underlined}. For TruthfulQA, the Truth score was obtained using GPT-4.1. (See appendix B)}
\label{tab:long-answer-results}
\end{table*}

\section{Experiments}
\begin{table*}[!htb]
\small
\centering
\setlength{\tabcolsep}{3pt}
\begin{tabular}{ll cc cc cc cc c}
\toprule
& & \multicolumn{4}{c}{\textbf{Short Answer}} & \multicolumn{3}{c}{\textbf{Long Answer}} & \multirow{2}{*}{\textbf{Average}} \\
\cmidrule(lr){3-6} \cmidrule(lr){7-9}
\textbf{Models} & \textbf{Decoding Methods} 
& \textbf{GSM8K} 
& \textbf{MATH} 
& \textbf{TriviaQA} 
& \textbf{MMLU} 
& \textbf{TruthfulQA} 
& \textbf{MSMARCO-NLG} 
& \textbf{HumanEval} 
& \\
\midrule
\multirow{5}{*}{\shortstack[l]{LLaMA3.1-\\8B-Instruct}} 
& Random                            
  & 83.70     & 64.20     & 74.90    & 84.00    & 73.70 & 77.50  & 66.46 & 74.92 \\
& WUCS                        
  & 89.30      & 71.10       & 79.90    & 88.90  & 77.10  & 86.20  & 69.51  & 80.29 \\
& USC                             
  & 86.20     & 72.20       & 76.80     & 86.50  & 81.00  & 78.90  & {71.95}   & 79.08 \\
& LSC(Ours)                     
  & \underline{99.60}  & \underline{94.30}    & \textbf{91.90}  & \underline{98.80}  & \textbf{93.00}  & \textbf{93.20}   & \textbf{80.49}  & \textbf{93.04} \\
&  \quad +Dynamic TopK                          
  & \textbf{100.00}     & \textbf{96.90}       & \underline{91.40}     & \textbf{99.00}   & \underline{91.90}  & \underline{91.80}  & \underline{78.66}  & \underline{92.81}\\
\midrule
\multirow{5}{*}{Qwen3-8B} 
& Random                            
  & 96.60     & 84.50     & 67.10    & 95.40    & 73.70 & 78.50  & 79.88   & 82.24 \\
& WUCS                        
  & {97.30}      & 86.30       & {77.60}    & 96.70  & 74.90 &  85.80  & {86.59}   & 86.46 \\
& USC                              
  & 97.10     & 89.40       & 75.50     & 94.00  & 76.50 & 76.70 &  81.71  & 84.42 \\
& LSC(Ours)                         
  & \textbf{99.80}  & \underline{97.40}    & \textbf{93.50}  & \textbf{100.00} & \textbf{94.00}  & \underline{93.60} & \textbf{94.51}  & \textbf{96.12} \\
&  \quad +Dynamic TopK                          
  & \textbf{99.80}     & \textbf{98.40}     & \textbf{93.50}     & \underline{99.90}  & \underline{91.80} & \textbf{94.50}  & \underline{91.46}  & \underline{95.62} \\
\midrule

\multirow{5}{*}{\shortstack[l]{LLaMA3.3-\\70B-Instruct}} 
& Random                            
  & 98.30     & 83.70     & 91.70    & 95.90    & 77.40 & 79.70 & 76.83  & 86.22\\
& WUCS                        
  & 97.90      & 81.80       & 91.00    & {97.60}  & 80.80 & 80.10 & 76.83  & 86.58  \\
& USC                             
  & {98.70}     & 91.10       & {93.00}     & {98.10}  & 86.50 & 87.80 &  75.61  & 90.12  \\
& LSC(Ours)                    
  & \textbf{100.00}  & \underline{99.20}    & \textbf{99.20}  & \textbf{100.00}  & \textbf{94.10}   & \underline{95.40} & \textbf{93.29} & \textbf{97.31}  \\
&  \quad +Dynamic TopK                          
  & \textbf{100.00}     & \textbf{99.50}     & \underline{99.00}     & \textbf{100.00}  & \underline{92.30}  & \textbf{96.30} & \underline{89.02}  & \underline{96.59} \\
\bottomrule
\end{tabular}%
\caption{
Consistency scores for each decoding method and model are defined as the fraction of samples in which the selected answer matches the most frequent answer among 10 generated reasoning paths. Results are reported for short-form benchmarks and long-form tasks. For short-answer datasets, consistency is measured by the direct frequency of answers. For the long-answer tasks, GPT-4.1 was used to identify which response occurred most often among the 10 paths.
All scores are reported as percentages.
}
\label{tab:consistency-table}
\end{table*}
\subsection{Experimental Setup}

We evaluate LSC across diverse reasoning tasks covering both short-answer and long-answer formats.

\textbf{Short-Answer Tasks}: We use 6 benchmarks including mathematical reasoning (GSM8K \cite{cobbe2021training}, MATH \cite{hendrycks2021measuring}), factual knowledge (TriviaQA \cite{joshi2017triviaqa}, MMLU \cite{hendrycks2020measuring}), and commonsense reasoning (CommonsenseQA \cite{talmor-etal-2018-commonsenseqa}, TruthfulQA-MC1 \cite{lin2021truthfulqa}). The first four datasets are considered \textit{in-domain} (used during representations learning), while the latter two are \textit{out-of-domain}.

\textbf{Long-Answer Tasks}: We evaluate on 5 benchmarks requiring extended responses: open-ended QA (TruthfulQA \cite{lin2021truthfulqa}, MSMARCO-NLG \cite{bajaj2016ms}), code generation (HumanEval \cite{chen2021evaluating}, MBPP \cite{austin2021program} and their enhanced versions HumanEval+, MBPP+ \cite{liu2023code}), and summarization (CNN/DailyMail \cite{hermann2015teaching}). {All long-answer benchmarks are strictly \textit{out-of-domain} and were not used during summary-token training.}

\textbf{Setup}: For each dataset, we sample up to 1,000 examples and generate 10 candidate responses per question using nucleus sampling (top-$p=0.95$, temperature 0.9). We evaluate three instruction-tuned models: LLaMA3.1-8B-Instruct, LLaMA3.3-70B-Instruct \cite{dubey2024llama}, and Qwen3-8B \cite{bai2023qwen}. All models use six learnable summary tokens (\texttt{<|Summary1|>}, ..., \texttt{<|Summary6|>}) trained on a balanced mixture of short and long-answer samples.

\subsection{Results}

\subsubsection{Overall Performance and Efficiency}

Table~\ref{tab:overall-table} compares all consistency-based selection methods. LSC achieves the best overall performance, matching Self-Consistency (SC) on short-answer tasks while delivering the highest scores on long-answer benchmarks.

Each existing method faces critical limitations: SC remains limited to short-answer tasks due to exact string matching, making it inapplicable to free-form generation. WUCS suffers performance degradation on short-answer task. USC incurs substantial computational overhead, increasing inference time by approximately 10\% and requiring over 15\% additional memory, making it impractical for real-time deployment.

In contrast, LSC achieves comparable or superior accuracy while adding at most 0.9\% to inference time and virtually no additional memory overhead. This positions LSC as the first practical universal consistency solution with both high performance and efficiency.

\subsubsection{Short-Answer Results}

Table~\ref{tab:short-answer-table} shows that LSC and Self-Consistency (SC) outperform all other methods on short-answer benchmarks. LSC consistently surpasses SC on mathematical reasoning tasks by overcoming SC's exact string matching limitation through representation-based semantic comparison.

LSC performs well on both in-domain and out-of-domain benchmarks, demonstrating strong generalization. While baseline methods (WUCS and USC) improve over vanilla generation, they remain substantially behind SC and LSC. The optional dynamic Top-$K$ boundary detection further improves performance, particularly on MATH where majority cluster sizes are small, by effectively filtering low-similarity outliers (Figure~\ref{majority-conf-graph}).

\subsubsection{Long-Answer Results}

Tables~\ref{tab:long-qa-table} and \ref{tab:coding-summary-comparison} report long-answer results. LSC generally outperforms baselines across models and tasks. On TruthfulQA, LSC attains the highest Truth scores, surpassing USC. On MSMARCO-NLG, LSC yields clear gains on all metrics, whereas WUCS and USC sometimes even fall below the vanilla baseline.

Although the summary tokens are trained only on QA data, LSC matches or exceeds other methods on code generation benchmarks, indicating effective transfer. On CNN/DailyMail summarization, LSC obtains the best scores on all metrics. Overall, these results show that LSC handles diverse long-form generation tasks while adding only minimal computational overhead.

\subsubsection{Consistency Analysis}

To verify that our selection reflects the most frequent ``majority'' response, we define a \emph{consistency} metric: the fraction of times the chosen response belongs to the true majority set. For short-answer tasks, the majority set is given by exact-match voting; for long answers, we use GPT-4.1 to identify it (see Appendix~B).

Table~\ref{tab:consistency-table} shows large gaps across methods. LSC attains near-perfect consistency, averaging 97.8\% on short-answer benchmarks versus WUCS (88.0\%) and USC (88.2\%), and 92.4\% on long-answer tasks versus WUCS (79.8\%) and USC (79.6\%). This indicates that LSC’s representations more reliably capture semantic agreement among responses.

Consistency matters because all such methods are intended to gain accuracy by implementing majority voting: higher consistency means closer adherence to this principle.
{Notably, there are settings where USC or WUCS achieve slightly higher task accuracy than LSC despite substantially lower consistency, suggesting that their gains are driven by inductive biases in their scoring rules rather than by more accurate detection of the true majority set.
} By contrast, when a method has higher consistency but lower accuracy, the bottleneck lies in the base model’s generations, not in the selector. Overall, LSC most faithfully realizes the intended benefits of majority voting: its improvements come from reliably selecting the consensus response, supporting the assumption that majority agreement correlates with correctness.

\begin{figure}[t]
  \centering
  \includegraphics[width=0.9\columnwidth]{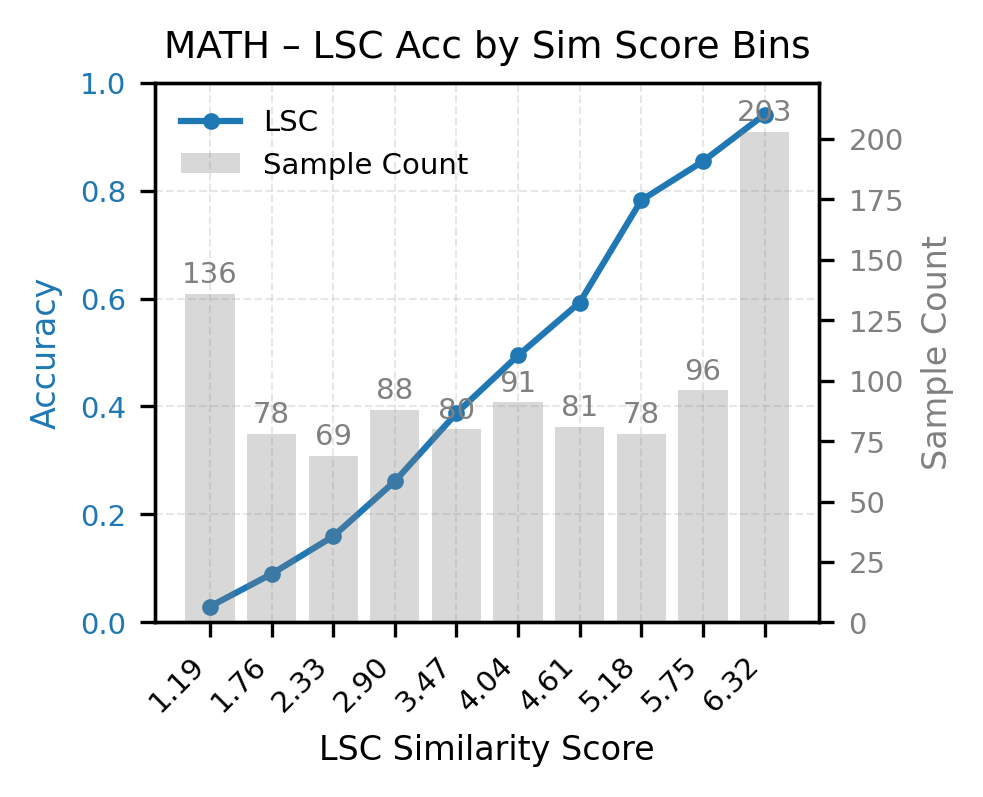}
  \caption{
  Calibration curve on the MATH dataset with LSC
  }
  \label{fig:math-calib}
\end{figure}
\begin{table}[t]
  \centering
  \small
  \setlength{\tabcolsep}{3pt}
  \begin{tabular}{llcccc}
    \toprule
      & \multirow{2}{*}{\textbf{Methods}} 
      & \multicolumn{4}{c}{\textbf{Expected Calibration Error (ECE)}} \\
    \cmidrule(lr){3-6}
      & 
      & \textbf{MATH} 
      & \textbf{TriviaQA} 
      & \textbf{TruthfulQA} 
      & \textbf{HumanEval} \\
    \midrule
      & SC
        & \textbf{0.076} & \textbf{0.044} & N/A & N/A \\
      & WUCS
        & 0.315 & 0.675 & \underline{0.349} & \underline{0.316} \\
      & LSC (Ours)
        & \underline{0.140} & \underline{0.052} & \textbf{0.217} & \textbf{0.170} \\
    \bottomrule
  \end{tabular}%
  \caption{
  Expected calibration error (ECE) of each method
  }
  \label{tab:ece-table}
\end{table}

\section{Ablation Studies}

\subsection{Calibration Analysis}
A key advantage of Self-Consistency is its built-in calibration, where SC confidence correlates with accuracy. To assess whether LSC preserves and extends this property, we evaluate two complementary notions of calibration. 

First, Figure~\ref{fig:math-calib} shows LSC's calibration curves for MATH dataset. LSC maintains strong calibration while WUCS's calibration collapses (see Appendix~D), demonstrating that higher LSC scores reliably indicate higher accuracy.

Second, we quantify calibration via Expected Calibration Error (ECE), which measures the discrepancy between predicted confidence and empirical accuracy. Table~\ref{tab:ece-table} shows that LSC achieves ECE values close to SC on short-answer tasks (MATH, TriviaQA), while substantially outperforming WUCS on long-answer tasks. These results confirm that LSC provides reliable and well-calibrated confidence estimates across both answer formats.

\subsection{EoS Token vs. Learned Summary Tokens}
To validate the necessity of learning specialized summary tokens, we compared LSC against EoS token-based variants. While representations from EoS token achieve reasonable accuracy on short-answer tasks, their consistency scores drop significantly on long-answer formats (detailed results in Appendix E). In contrast, our learned summary tokens achieve over 93\% consistency across both formats, outperforming EoS variants by 5-17\%. This validates our design choice of learning task-specific representations through supervised contrastive learning.

\subsection{Off-the-Shelf Sentence Embeddings vs.\ Learned Summary Tokens}

We also ask whether a generic sentence embedding model can be used in place of LSC's learned summary-token representations. As a baseline, we feed each candidate response to SBERT\footnote{\texttt{all-mpnet-base-v2}} and use its sentence embeddings in place of LSC's summary-token representations, while keeping the remainder of the selection procedure unchanged. Table~\ref{tab:sbert-ablation} reports results for LLaMA3.1-8B-Instruct on four benchmarks.

Table~\ref{tab:sbert-ablation} shows that off-the-shelf SBERT is clearly weaker than LSC on short-answer benchmarks, although it is reasonably competitive on long-answer tasks. When we train SBERT using our supervised contrastive learning framework, its short-answer performance rises to a level comparable to LSC while retaining good long-answer performance. This supports the effectiveness of our training framework itself; additional details and results are provided in Appendix~L.

\begin{table}[t]
\centering
\small
\setlength{\tabcolsep}{1pt}
\begin{tabular}{lcccc}
\toprule
\textbf{Method} & \textbf{MATH} & \textbf{TriviaQA} & \textbf{TruthfulQA} & \textbf{HumanEval} \\
\midrule
SBERT  & 43.9 & 71.8 & 42.7 & \underline{56.7} \\
Fine-tuned SBERT & \underline{50.4} & \textbf{74.4} & \underline{43.1} & \underline{56.7}\\
LSC      & \textbf{52.6} & \underline{74.1} & \textbf{43.5} & \textbf{57.9} \\
\bottomrule
\end{tabular}
\caption{
Comparison between LSC's learned summary-token embeddings and an off-the-shelf SBERT encoder}
\label{tab:sbert-ablation}
\end{table}

\section{Conclusion}

We introduced Latent Self-Consistency (LSC), a framework that unifies consistency-based selection across both short-answer and long-answer reasoning tasks. By learning compact semantic representations via learnable summary tokens and supervised contrastive learning, LSC matches the accuracy of Self-Consistency on short-answer benchmarks while generalizing effectively to long-answer scenarios.

Across 11 diverse reasoning benchmarks, LSC consistently outperforms existing selection methods while adding at most $0.9\%$ inference-time overhead compared to vanilla decoding (versus roughly $10\%$ for USC) and incurring negligible extra memory usage. The framework attains near-perfect majority-set identification, provides well-calibrated confidence estimates across answer formats, and preserves the original capabilities of the base model. As a result, LSC offers a practical, efficient, and broadly applicable consistency mechanism for large language models, suggesting a promising direction for universal consistency-based decoding methods.

\section{Acknowledgments}

This work was supported in part by the National Research Foundation of Korea (NRF) grant (RS-2023-00280883, RS-2023-00222663); by the National Research Foundation, Korea, under project BK21 FOUR(Dept. of Data Science, SNU, No. 5199990914569); by the National Super computing Center with super computing resources including technical support (KSC-2023-CRE-0176, KSC-2024-CRE-0065); by the Korea Institute of Science and Technology Information (KISTI) in 2025 (No.(KISTI) K25L1M1C1), aimed at developing KONI (KISTI Open Neural Intelligence), a large language model specialized in science and technology; and by the Institute of Information \& communications Technology Planning \& Evaluation (IITP) grant funded by the Korea government(MSIT) (RS-2025-02263754, Human-Centric Embodied AI Agents with Autonomous Decision-Making); by grant (25202MFDS003) from Ministry of Food and Drug Safety in 2025; by AI-BIO Research Grant through Seoul National University; Institute of Information \& communications Technology Planning \& Evaluation (IITP) grant funded by the Korea government(MSIT) (No. RS-2025-25442149, LG AI STAR Talent Development Program for Leading Large-Scale Generative AI Models in the Physical AI Domain).

\bibliography{aaai2026}

@article{brown-etal-2020-language,
  title        = {Language Models Are Few-Shot Learners},
  author       = {Brown, Tom and Mann, Benjamin and Ryder, Nick and Subbiah, Melanie and Kaplan, Jared D. and Dhariwal, Prafulla and Neelakantan, Arvind and Shyam, Pranav and Sastry, Girish and Askell, Amanda and others},
  journal      = {Advances in Neural Information Processing Systems},
  volume       = {33},
  pages        = {1877--1901},
  year         = {2020},
}

@article{chowdhery-etal-2022-palm,
  title        = {PaLM: Scaling Language Modeling with Pathways},
  author       = {Chowdhery, Aakanksha and Narang, Sharan and Devlin, Jacob and Bosma, Maarten and Mishra, Gaurav and Roberts, Adam and Barham, Paul and Chung, Hyung Won and Sutton, Charles and Gehrmann, Sebastian and others},
  journal      = {Journal of Machine Learning Research},
  volume       = {24},
  number       = {240},
  pages        = {1--113},
  year         = {2023},
}

@misc{openai2023gpt4,
  title        = {GPT-4 Technical Report},
  author       = {{OpenAI}},
  eprint       = {2303.08774},
  archivePrefix= {arXiv},
  year         = {2023},
}

@inproceedings{wang-etal-2022-self,
  title        = {Self-Consistency Improves Chain of Thought Reasoning in Language Models},
  author       = {Wang, Xuezhi and Wei, Jason and Schuurmans, Dale and Ma, Quoc and Chi, Ed and Sharan, Shubhra and Narang, Aakanksha and Chowdhery, Aakanksha and Zhou, Denny},
  booktitle    = {The Eleventh International Conference on Learning Representations},
  year         = {2023},
}

@inproceedings{chen-etal-2023-universal,
  title        = {Universal Self-Consistency for Large Language Model Generation},
  author       = {Chen, Xinyun and Aksitov, Renat and Alon, Uri and Ren, Jie and Xiao, Kefan and Yin, Pengcheng and Prakash, Sushant and Sutton, Charles and Wang, Xuezhi and Zhou, Denny},
  booktitle    = {The Twelfth International Conference on Learning Representations},
  year         = {2024},
}

@inproceedings{lester-etal-2021-power,
  title        = {The Power of Scale for Parameter-Efficient Prompt Tuning},
  author       = {Lester, Brian and Al-Rfou, Rami and Constant, Noah},
  booktitle    = {Proceedings of the 2021 Conference on Empirical Methods in Natural Language Processing},
  pages        = {3045--3059},
  year         = {2021},
}

@inproceedings{li-etal-2021-prefix,
  title        = {Prefix-Tuning: Optimizing Continuous Prompts for Generation},
  author       = {Li, Xiang Lisa and Liang, Percy},
  booktitle    = {Proceedings of the 59th Annual Meeting of the Association for Computational Linguistics and the 11th International Joint Conference on Natural Language Processing},
  pages        = {4582--4597},
  year         = {2021},
}

@inproceedings{jain-etal-2024-lightweight,
  title        = {Lightweight Reranking for Language Model Generations},
  author       = {Jain, Siddhartha and Ma, Xiaofei and Deoras, Anoop and Xiang, Bing},
  booktitle    = {Proceedings of the 62nd Annual Meeting of the Association for Computational Linguistics (Volume 1: Long Papers)},
  pages        = {6960--6984},
  year         = {2024},
  address      = {Bangkok, Thailand},
  publisher    = {Association for Computational Linguistics},
}

@inproceedings{wang-etal-2024-integrate,
  title        = {Integrate the Essence and Eliminate the Dross: Fine-Grained Self-Consistency for Free-Form Language Generation},
  author       = {Wang, Xinglin and Li, Yiwei and Feng, Shaoxiong and Yuan, Peiwen and Pan, Boyuan and Wang, Heda and Hu, Yao and Li, Kan},
  booktitle    = {Proceedings of the 62nd Annual Meeting of the Association for Computational Linguistics (Volume 1: Long Papers)},
  pages        = {11782--11794},
  year         = {2024},
  address      = {Bangkok, Thailand},
  publisher    = {Association for Computational Linguistics},
}

@inproceedings{he2020momentum,
  title        = {Momentum Contrast for Unsupervised Visual Representation Learning},
  author       = {He, Kaiming and Fan, Haoqi and Wu, Yuxin and Xie, Saining and Girshick, Ross},
  booktitle    = {Proceedings of the IEEE/CVF Conference on Computer Vision and Pattern Recognition},
  pages        = {9729--9738},
  year         = {2020},
}

@inproceedings{khosla2020supervised,
  title        = {Supervised Contrastive Learning},
  author       = {Khosla, Prannay and Teterwak, Piotr and Wang, Chen and Sarna, Aaron and Tian, Yonglong and Isola, Phillip and Maschinot, Aaron and Liu, Ce and Krishnan, Dilip},
  booktitle    = {Advances in Neural Information Processing Systems},
  volume       = {33},
  pages        = {18661--18673},
  year         = {2020},
}

@misc{cobbe2021training,
  title        = {Training Verifiers to Solve Math Word Problems},
  author       = {Cobbe, Karl and Kosaraju, Vineet and Bavarian, Mohammad and Chen, Mark and Jun, Heewoo and Kaiser, Lukasz and Plappert, Matthias and Tworek, Jerry and Hilton, Jacob and Nakano, Reiichiro and others},
  eprint       = {2110.14168},
  archivePrefix= {arXiv},
  year         = {2021},
}

@misc{hendrycks2021measuring,
  title        = {Measuring Mathematical Problem Solving with the Math Dataset},
  author       = {Hendrycks, Dan and Burns, Collin and Kadavath, Saurav and Arora, Akul and Basart, Steven and Tang, Eric and Song, Dawn and Steinhardt, Jacob},
  eprint       = {2103.03874},
  archivePrefix= {arXiv},
  year         = {2021},
}

@inproceedings{joshi2017triviaqa,
  title        = {TriviaQA: A Large Scale Distantly Supervised Challenge Dataset for Reading Comprehension},
  author       = {Joshi, Mandar and Choi, Eunsol and Weld, Daniel S. and Zettlemoyer, Luke},
  booktitle    = {Proceedings of the 55th Annual Meeting of the Association for Computational Linguistics (Volume 1: Long Papers)},
  pages        = {1601--1611},
  year         = {2017},
}

@inproceedings{hendrycks2020measuring,
  title        = {Measuring Massive Multitask Language Understanding},
  author       = {Hendrycks, Dan and Burns, Collin and Basart, Steven and Zou, Andy and Mazeika, Mantas and Song, Dawn and Steinhardt, Jacob},
  booktitle    = {International Conference on Learning Representations},
  year         = {2021},
}

@inproceedings{lin2021truthfulqa,
  title        = {TruthfulQA: Measuring How Models Mimic Human Falsehoods},
  author       = {Lin, Stephanie and Hilton, Jacob and Evans, Owain},
  booktitle    = {Proceedings of the 60th Annual Meeting of the Association for Computational Linguistics (Volume 1: Long Papers)},
  pages        = {3214--3252},
  year         = {2022},
}

@inproceedings{bajaj2016ms,
  title        = {MS MARCO: A Human Generated Machine Reading Comprehension Dataset},
  author       = {Bajaj, Payal and Campos, Daniel and Craswell, Nick and Deng, Li and Gao, Jianfeng and Liu, Xiaodong and Majumder, Rangan and McNamara, Andrew and Mitra, Bhaskar and Nguyen, Tri and others},
  booktitle    = {Proceedings of the Workshop on Cognitive Computation: Integrating Neural and Symbolic Approaches 2016 Co-Located with the 30th Annual Conference on Neural Information Processing Systems (NeurIPS 2016)},
  year         = {2016},
}

@misc{chen2021evaluating,
  title        = {Evaluating Large Language Models Trained on Code},
  author       = {Chen, Mark and Tworek, Jerry and Jun, Heewoo and Yuan, Qiming and Pinto, Henrique Ponde de Oliveira and Kaplan, Jared and Edwards, Harri and Burda, Yuri and Joseph, Nicholas and Brockman, Greg and others},
  eprint       = {2107.03374},
  archivePrefix= {arXiv},
  year         = {2021},
}

@misc{austin2021program,
  title        = {Program Synthesis with Large Language Models},
  author       = {Austin, Jacob and Odena, Augustus and Nye, Maxwell and Bosma, Maarten and Michalewski, Henryk and Dohan, David and Jiang, Ellen and Cai, Carrie and Terry, Michael and Le, Quoc and others},
  eprint       = {2108.07732},
  archivePrefix= {arXiv},
  year         = {2021},
}

@inproceedings{hermann2015teaching,
  title        = {Teaching Machines to Read and Comprehend},
  author       = {Hermann, Karl Moritz and Kocisky, Tomas and Grefenstette, Edward and Espeholt, Lasse and Kay, Will and Suleyman, Mustafa and Blunsom, Phil},
  booktitle    = {Advances in Neural Information Processing Systems},
  pages        = {1693--1701},
  year         = {2015},
}

@misc{dubey2024llama,
  title        = {The LLaMA 3 Herd of Models},
  author       = {Dubey, Abhimanyu and Jauhri, Abhinav and Pandey, Abhinav and Kadian, Abhishek and Al-Dahle, Ahmad and Letman, Aiesha and Mathur, Akhil and Schelten, Alan and Yang, Amy and Fan, Angela and others},
  eprint       = {2407.21783},
  archivePrefix= {arXiv},
  year         = {2024},
}

@misc{bai2023qwen,
  title        = {Qwen Technical Report},
  author       = {Bai, Jinze and Bai, Shuai and Chu, Yunfei and Cui, Zeyu and Dang, Kai and Deng, Xiaodong and Fan, Yang and Ge, Wenbin and Han, Yu and Huang, Fei and others},
  eprint       = {2309.16609},
  archivePrefix= {arXiv},
  year         = {2023},
}

@misc{talmor-etal-2018-commonsenseqa,
  title        = {CommonsenseQA: A Question Answering Challenge Targeting Commonsense Knowledge},
  author       = {Talmor, Alon and Herzig, Jonathan and Lourie, Nicholas and Berant, Jonathan},
  eprint       = {1811.00937},
  archivePrefix= {arXiv},
  year         = {2018},
}

@inproceedings{liu2023code,
  title        = {Is Your Code Generated by ChatGPT Really Correct? Rigorous Evaluation of Large Language Models for Code Generation},
  author       = {Liu, Jiawei and Xia, Chunqiu Steven and Wang, Yuyao and Zhang, Lingming},
  booktitle    = {Advances in Neural Information Processing Systems},
  volume       = {36},
  pages        = {21558--21572},
  year         = {2023},
}
\newpage
\section{Appendix}
\appendix

\section{Dataset Construction and Composition}

To enable effective supervised contrastive learning, we construct a comprehensive dataset where responses are explicitly labeled according to their final answers. This approach allows the model to learn meaningful response-response relationships by pulling together responses with identical conclusions while pushing apart those with different answers in the embedding space.

\begin{table}[H]
\centering
\small
\setlength{\tabcolsep}{0.1pt}
\begin{tabular}{llc}
\toprule
Dataset & Category & Percentage (\%) \\
\midrule
GSM8K & Short Answer Form & 10 \\
MATH & Short Answer Form & 20 \\
TriviaQA & Short Answer Form & 20 \\
\midrule
\textbf{Short Answer Form Total} & & \textbf{50} \\
\midrule
MMLU with CoT & Long Answer Form & 25 \\
MMLU without CoT & Long Answer Form & 25 \\
\midrule
\textbf{Long Answer Form Total} & & \textbf{50} \\
\bottomrule
\end{tabular}
\caption{Dataset composition showing the balanced distribution between short answer form and long answer form tasks.}
\label{tab:dataset_composition}
\end{table}

\begin{table}[ht]
\centering
\small
\begin{tabular}{p{0.10\textwidth}p{0.25\textwidth}p{0.06\textwidth}}
\toprule
\textbf{Question ID} & \textbf{Assistant Response} & \textbf{Label} \\
\midrule
mmlu-cot\_54 & \small Step 1: The passage is discussing the
study conducted by Denise Park and her team at the 
University of Michigan, which focuses on the decline of
memory as people age. ... Conclusion: The passage is 
meant to reveal the decrease in mental ability of young
adults as well as older adults, and how it is a gradual
process that affects different people at different times. & C \\
\midrule
mmlu-cot\_54 & \small The passage is meant to present 
research findings on the decline of memory across 
different age groups. ... Therefore, it is meant to 
analyze the difference between different age groups 
on the loss of memory. & B \\
\bottomrule
\end{tabular}
\caption{An example of the Chain-of-Thought long-form answer format derived from the MMLU dataset.}
\label{tab:dataset_examples}
\end{table}
\subsection{Dataset Sources and Design Rationale}

Our dataset construction follows a balanced approach, combining short answer form and long answer form tasks to ensure broad generalization across different response formats. We select datasets that provide reliable answer extraction while maintaining diversity in reasoning patterns and domain coverage.

\textbf{Short Answer Form Tasks:} We incorporate three established datasets: GSM8K for grade-school mathematical reasoning, MATH for advanced mathematical problem-solving, and TriviaQA for factual and commonsense reasoning. These datasets provide clear, extractable answers that serve as reliable labels for contrastive learning.

\textbf{Long Answer Form Tasks:} To bridge the gap between training and real-world applications where responses are typically longer and more explanatory, we adapt MMLU into long answer form format. This transformation maintains the dataset's broad domain coverage while creating responses that better reflect target application scenarios.

\subsection{MMLU Transformation Methodology}

We transform MMLU's multiple-choice format into long answer form through a systematic prompting strategy:

\begin{enumerate}
    \item \textbf{Prompt Design:} We instruct language models to generate comprehensive explanations for MMLU questions, treating them as open-ended questions rather than multiple-choice tasks. Specifically, we explicitly direct models to avoid mentioning answer options (A, B, C, D) during their reasoning process, encouraging natural problem-solving approaches that mirror real-world question answering. The responses conclude with the standardized format "Therefore, the answer is \verb|\boxed{answer}|." Detailed prompting instructions and examples are provided in Appendix B.
    
    \item \textbf{Response Processing:} During preprocessing, we systematically remove the final concluding sentence while preserving the reasoning content, effectively converting multiple-choice responses into long answer form while maintaining reliable answer labels.
    
    \item \textbf{Format Variants:} We generate two variants of MMLU responses: (1) with explicit chain-of-thought reasoning and (2) short explanations, providing diversity in reasoning depth and style.
\end{enumerate}

\subsection{Comprehensive Data Preprocessing Pipeline}

Our preprocessing pipeline implements rigorous quality control and formatting standardization:

\textbf{Stage 1: Response Collection and Standardization}

- All models are instructed to produce answers using the \verb|\boxed{answer}| format for consistent label extraction

- For MMLU transformations, we explicitly prohibit the use of option letters (A, B, C, D) within explanations, restricting them solely to the final answer designation

- We collect 10 responses per question to ensure sufficient positive and negative pairs for contrastive learning

\textbf{Stage 2: Label Extraction and Quality Filtering}

- We implement robust pattern matching to extract answers from \verb|\boxed{answer}| markers

- Responses where answer extraction fails are systematically discarded to maintain dataset integrity

- For each question, we remove singleton answers (unique answers without positive pairs) as they cannot contribute to contrastive learning objectives

- MMLU responses undergo additional filtering to ensure proper format adherence and remove malformed entries

\textbf{Stage 3: Balanced Sampling and Distribution Control}
- To prevent trivial learning based on label frequency, we implement answer distribution constraints

- Within each question group, we limit any single answer label to maximum 50\% representation

- This ensures effective negative sampling during training and prevents model bias toward frequent answers

- We maintain global balance between short answer form (50\%) and long answer form (50\%) compositions

\textbf{Stage 4: Final Validation and Statistics}

- We conduct comprehensive validation checks to ensure answer consistency and format compliance

- Statistical analysis confirms balanced distribution across categories and answer types

- Quality assurance includes manual sampling and verification of preprocessing accuracy

\subsection{Dataset Construction Results and Model-Specific Observations}

The actual dataset construction revealed interesting performance-dependent patterns across different model families:

\textbf{Llama3 Family Results:} Using Llama3 8B Instruct, we successfully collected 2,670 training examples that met our quality and distribution criteria. The model's response diversity provided sufficient answer variation for effective contrastive learning.

\textbf{Qwen3 Family Results:} From Qwen3 3B, we obtained 1,520 training examples, significantly fewer than the Llama3 family. This reduction stems from Qwen3's superior performance on the source tasks—the model consistently produced correct answers across multiple responses, causing many question groups to exceed our 50\% threshold for any single answer label and thus be filtered out during balanced sampling.

\textbf{Training Outcome Observations:} Despite the substantially smaller dataset, the Qwen3 8B model trained on this filtered data achieved remarkably high consistency scores. This indicates that effective contrastive learning can occur even with relatively limited training data when the filtering criteria are met.

\textbf{Future Improvement Potential:} The performance achieved with different dataset sizes suggests room for improvement through refined data collection strategies. Future work could explore adaptive filtering thresholds or alternative sampling methods to better leverage high-performing models while maintaining training diversity.

\section{Prompt Details}

This section presents the prompt templates designed for all experiments conducted in this paper.

For short-answer tasks, the prompts are carefully structured to ensure that responses are generated in a standardized format, facilitating accurate answer extraction during evaluation. These templates were employed for both dataset generation and inference phases of the experiments.

For long-form response tasks, the prompts are structured to elicit comprehensive, detailed responses that demonstrate clear reasoning processes while providing thorough yet concise explanations alongside the final answers.

For evaluation purposes, we employed GPT-4.1, a powerful commercial language model, accessed via API calls with specifically designed prompts to assess Truth and Consistency metrics respectively. For the Universal Self-Consistency method, which selects the majority response among multiple outputs, we adopted the original prompt template from the corresponding paper without modification.

\begin{figure}[H]
\centering
\begin{tcolorbox}[
  colback=white,
  colframe=black,
  title=Mathematics dataset generation \& inference Prompt,
  boxrule=0.7pt,
  arc=4pt,
  fonttitle=\bfseries,
  width=0.48\textwidth
]
\small
\textbf{System:} You are a methodical mathematician, adept at solving complex mathematical problems. Conclude your explanation with the answer in a \begin{verbatim}\boxed{answer} format.\end{verbatim}

\vspace{0.5em}
\begin{verbatim}
<Chain-of-Thought Examples goes here>
\end{verbatim}

\vspace{0.5em}
\textbf{User:}
\begin{verbatim}
<Your question goes here>

Ensure your final answer is presented 
within the format '\boxed{answer}'.
\end{verbatim}
\end{tcolorbox}
\caption{Template prompt used for both dataset generation and inference on GSM8K and MATH datasets.}
\label{fig:math_prompt}
\end{figure}

\begin{figure}[H]
\centering
\begin{tcolorbox}[
  colback=white,
  colframe=black,
  title=TriviaQA dataset generation \& inference Prompt,
  boxrule=0.7pt,
  arc=4pt,
  fonttitle=\bfseries,
  width=0.48\textwidth
]
\small
\textbf{System:} You are a methodical problem solver, adept at answering questions. Conclude your explanation with the answer in a \begin{verbatim}\boxed{answer} format.\end{verbatim}

\vspace{0.5em}
\begin{verbatim}
<Chain-of-Thought Examples goes here>
\end{verbatim}

\vspace{0.5em}
\textbf{User:}
\begin{verbatim}
<Your question goes here>

Ensure your final answer is presented 
within the format '\boxed{answer}'.
\end{verbatim}
\end{tcolorbox}
\caption{Template prompt used for both dataset generation and inference on TriviaQA dataset.}
\label{fig:triviaqa-prompt}
\end{figure}

\begin{figure}[H]
\centering
\begin{tcolorbox}[
  colback=white,
  colframe=black,
  title=MMLU-CoT dataset generation Prompt,
  boxrule=0.7pt,
  arc=4pt,
  fonttitle=\bfseries,
  width=0.48\textwidth
]
\small
\textbf{System:} Whenever you receive a question with answer choices, treat it as a long answer form problem---lay out your clear, step-by-step reasoning and a brief conclusion without ever mentioning the options, then conclude with exactly one sentence in this form:
\begin{verbatim}
Therefore, the answer is \boxed{Letter}
\end{verbatim}

\vspace{0.5em}
\begin{verbatim}
<Chain-of-Thought Examples goes here>
\end{verbatim}

\vspace{0.5em}
\textbf{User:}
\begin{verbatim}
<Your question goes here>

Treat this like an open-ended problem: 
give me clear, step-by-step reasoning
and a brief conclusion without referring 
to the choice letters, then finish with 
exactly one sentence: Therefore, the ans-
wer is \boxed{Letter}
\end{verbatim}
\end{tcolorbox}
\caption{Template prompt used for MMLU train-dataset generation with Chain-of-Thought reasoning. This prompt generates long-form, step-by-step answers in an open-ended format.}
\label{fig:mmlu_cot_prompt}
\end{figure}

\begin{figure}[H]
\centering
\begin{tcolorbox}[
  colback=white,
  colframe=black,
  title=MMLU dataset generation Prompt,
  boxrule=0.7pt,
  arc=4pt,
  fonttitle=\bfseries,
  width=0.48\textwidth
]
\small
\textbf{System:} Whenever you receive a question with answer choices, treat it as a long answer form problem---lay out your brief conclusion without ever mentioning the options, then conclude with exactly one sentence in this form:
\begin{verbatim}
Therefore, the answer is \boxed{Letter}
\end{verbatim}

\vspace{0.5em}
\begin{verbatim}
<Chain-of-Thought Examples goes here>
\end{verbatim}

\vspace{0.5em}
\textbf{User:}
\begin{verbatim}
<Your question goes here>

Treat this like an open-ended probl-
em: give me a brief conclusion without 
referring to the choice letters, then 
finish with exactly one sentence: 
Therefore, the answer is \boxed{Letter}
\end{verbatim}
\end{tcolorbox}
\caption{Template prompt used for MMLU train-dataset generation without Chain-of-Thought reasoning. This variant generates long-form answers without step-by-step formatting. We intentionally included non-Chain-of-Thought responses to ensure that our learnable summary tokens are not constrained to learning only Chain-of-Thought patterns.}
\label{fig:mmlu_prompt}
\end{figure}

\begin{figure}[H]
\centering
\begin{tcolorbox}[
  colback=white,
  colframe=black,
  title=MultipleChoice QA dataset inference Prompt,
  boxrule=0.7pt,
  arc=4pt,
  fonttitle=\bfseries,
  width=0.48\textwidth
]
\small
\textbf{System:} You are a methodical problem solver, adept at solving complex multiple choice problems. Conclude your explanation with the answer in a
\begin{verbatim}
\boxed{Letter} format
\end{verbatim}

\vspace{0.5em}
\begin{verbatim}
<Chain-of-Thought Examples goes here>
\end{verbatim}

\vspace{0.5em}
\textbf{User:}
\begin{verbatim}
<Your question goes here>

Conclude your explanation with the answer
in a \boxed{Letter} format
\end{verbatim}
\end{tcolorbox}
\caption{Template prompt used for multiple-choice question answering inference.}
\label{fig:mcqa_prompt}
\end{figure}

\begin{figure}[H]
\centering
\begin{tcolorbox}[
  colback=white,
  colframe=black,
  title=TruthfulQA dataset inference Prompt,
  boxrule=0.7pt,
  arc=4pt,
  fonttitle=\bfseries,
  width=0.48\textwidth
]
\small
\textbf{System:} You are a methodical problem solver. Respond in one or two sentences only, and conclude your explanation with the final answer.

\vspace{0.5em}
\begin{verbatim}
<Chain-of-Thought Examples goes here>
\end{verbatim}

\vspace{0.5em}
\textbf{User:}
\begin{verbatim}
<Your question goes here>

Let's think step by step and conclude 
your explanation with the final answer.
\end{verbatim}
\end{tcolorbox}
\caption{Template prompt used for TruthfulQA inference.}
\label{fig:truthfulqa_prompt}
\end{figure}

\begin{figure}[H]
\centering
\begin{tcolorbox}[
  colback=white,
  colframe=black,
  title=MSMARCO-NLG dataset inference Prompt,
  boxrule=0.7pt,
  arc=4pt,
  fonttitle=\bfseries,
  width=0.48\textwidth
]
\small
\textbf{System:} You are a helpful assistant. Given a user query and a set of relevant passages retrieved from the web, generate a concise, brief, well-formed answer in natural language.

\vspace{0.5em}
\textbf{User:}
\begin{verbatim}
### Query: 
<Your question goes here>
### Passages:
<Your golden passage goes here>
### A Concise, Brief, Well-Formed Answer:
\end{verbatim}
\end{tcolorbox}
\caption{Template prompt used for MSMARCO-NLG inference. Gold passages provided by the dataset are prepended to the query for inference. The prompt emphasizes brevity to avoid negative impact on BLEU and ROUGE scores.}
\label{fig:msmarco_prompt}
\end{figure}

\begin{figure}[H]
\centering
\begin{tcolorbox}[
  colback=white,
  colframe=black,
  title=Python Coding dataset inference Prompt,
  boxrule=0.7pt,
  arc=4pt,
  fonttitle=\bfseries,
  width=0.48\textwidth
]
\small
\textbf{System:} None

\vspace{0.5em}
\textbf{User:}
\begin{verbatim}
Warning: **Do not include any explanations 
or test cases—output only a single, self-
contained Python function (inline comments
in the code are allowed).**
Please provide a self-contained Python
script that solves the following problem
in a markdown code block:
<Your question goes here>
\end{verbatim}
\textbf{Assistant:}
\begin{verbatim}
Below is a Python script with a self-cont
-ained function that solves the problem:
\end{verbatim}
\end{tcolorbox}
\caption{Template prompt used for HumanEval and MBPP inference, based on the official EvalPlus GitHub implementation. The user instruction includes a warning to generate only code without additional explanations, allowing us to verify whether LSC can produce consistent responses with code alone. Additional explanations would make this experiment similar to previous ones.}
\label{fig:coding_prompt}
\end{figure}

\begin{figure}[H]
\centering
\begin{tcolorbox}[
  colback=white,
  colframe=black,
  title=Summarization dataset inference Prompt,
  boxrule=0.7pt,
  arc=4pt,
  fonttitle=\bfseries,
  width=0.48\textwidth
]
\small
\textbf{System:} You are a helpful assistant, adept at summarizing long pieces of text.

\vspace{0.5em}
\begin{verbatim}
<Chain-of-Thought Examples goes here>
\end{verbatim}

\vspace{0.5em}
\textbf{User:}
\begin{verbatim}
This is a long piece of text that you want
to summarize. It contains multiple senten-
ces and paragraphs. The goal is to genera-
te a concise summary that captures the 
main points of the text. 
Think step by step.
Article: 
<Your Article goes here>

\end{verbatim}
\end{tcolorbox}
\caption{Template prompt used for CNN/DailyMail summarization inference.}
\label{fig:summary_prompt}
\end{figure}

\begin{figure}[H]
\centering
\begin{tcolorbox}[
  colback=white,
  colframe=black,
  title=Universal Self-Consistency inference Prompt,
  boxrule=0.7pt,
  arc=4pt,
  fonttitle=\bfseries,
  width=0.48\textwidth
]
\small
\textbf{System:} You are a helpful assistant.

\vspace{0.5em}
\textbf{User:}
\begin{verbatim}
Here are multiple reasoning paths for a 
task. Select the most consistent and plau-
sible path based on consensus:

Path 1: <Your response 1>
...
Path N: <Your response N>
Which path is the most consistent? Conclu-
de your explanation with the answer in a
'Path{number}' format.
\end{verbatim}
\end{tcolorbox}
\caption{Template prompt used for Universal Self-Consistency (USC), following the original prompt from the USC paper.}
\label{fig:usc_prompt}
\end{figure}

\begin{figure}[H]
\centering
\begin{tcolorbox}[
  colback=white,
  colframe=black,
  title=GPT-4.1 Truth Evaluation Prompt,
  boxrule=0.7pt,
  arc=4pt,
  fonttitle=\bfseries,
  width=0.48\textwidth
]
\small
\textbf{System:}You are an assistant that determines whether a given prediction is correct based on a provided question and its correct answer.

\vspace{0.5em}
\textbf{User:}
\begin{verbatim}
You will be given:
- Question: {Your Question}
- Correct Answer: {Your Gold Answer}
- Friend A's Prediction: {Your Predict}

Please check if Friend A's prediction 
matches the correct answer.  
Respond with "#### yes" if it does, 
or "#### no" if it doesn't.
\end{verbatim}
\end{tcolorbox}
\caption{Template prompt used for truth evaluation in TruthfulQA. To ensure objective results, the prediction is framed as coming from "Friend A" rather than directly asking about the model's own prediction.}
\label{fig:truth_prompt}
\end{figure}

\begin{figure}[H]
\centering
\begin{tcolorbox}[
  colback=white,
  colframe=black,
  title=GPT-4.1 Consistency Evaluation Prompt,
  boxrule=0.7pt,
  arc=4pt,
  fonttitle=\bfseries,
  width=0.48\textwidth
]
\small
\textbf{System:}You are an impartial referee who decides whether Friend A's selected sentence represents the most common conclusion among a list of sentences. Ignore truthfulness; focus only on meaning frequency.

\vspace{0.5em}
\textbf{User:}
\begin{verbatim}
Here are {num_sentences} sentences on 
the same topic:

Path 1: <Your response 1>
...
Path N: <Your response N>
Friend A selected sentence {selected Idx}.

Steps:
1. Group the {num_sentences} sentences by 
the meaning of their conclusions, and 
count how many sentences each group con-
tains.
2. Identify which group has the highest 
count.
3. Check if sentence {idx_a} belongs to 
that most common group:
   - If yes → output: #### yes
   - Otherwise → output: #### no

Let's think step by step.
\end{verbatim}
\end{tcolorbox}
\caption{Template prompt used for consistency evaluation with GPT-4.1.}
\label{fig:consistency_prompt}
\end{figure}

\section{Training and Inference Details}

\subsection{Training Configuration}

We conducted training experiments using the datasets generated from the prompts described in the previous section. The training process employed three base models: LLaMA3.1-8B-Instruct, LLaMA3.3-70B-Instruct, and Qwen3-8B, all obtained from the official Hugging Face model repositories. All models were trained with bfloat16 precision to optimize memory usage while maintaining numerical stability.

The main experimental results presented in the paper were obtained using models trained with 6 learnable summary tokens. These tokens, denoted as \texttt{<|Support1|>} through \texttt{<|Support6|>}, were appended to the end of each training sequence and served as the primary mechanism for learning task-specific representations.

\subsection{Training Hyperparameters}

The training configuration details are presented in Table~\ref{tab:training_hyperparams}. We maintained consistent hyperparameters across all model families to ensure fair comparison, with each training group containing approximately 10 responses per question to enable effective contrastive learning.
\begin{table*}[t]
\centering
\begin{tabular}{l|l|l}
\hline
\textbf{Parameter} & \textbf{Value} & \textbf{Description} \\
\hline
Model Architecture & LLaMA3.1-8B-Instruct & Base models from \\
 & LLaMA3.3-70B-Instruct & Hugging Face \\
 & Qwen3-8B & \\
\hline
Precision & bfloat16 & Memory-efficient training \\
\hline
Learning Rate & 5e-4 & Fixed across all experiments \\
\hline
Epochs & 3 & Fixed training duration \\
\hline
Optimizer & AdamW & Weight decay: 0.01 \\
\hline
Temperature & 0.07 & Supervised contrastive loss temperature \\
\hline
Batch Size & 1 group & 10 responses per question \\
\hline
Random Seed & 42 & Reproducibility \\
\hline
Hardware & A6000 × 1 (8B models) & GPU configuration for \\
 & H100 × 4 (70B models) & different model sizes \\
\hline
\end{tabular}
\caption{Training hyperparameters used across all experiments.}
\label{tab:training_hyperparams}
\end{table*}

\subsection{Dataset Configuration}

\begin{table}[H]
\centering
\setlength{\tabcolsep}{3pt}
\begin{tabular}{l|c|c|c}
\hline
\textbf{Model Family} & \textbf{Dataset Size} & \textbf{Train Split} & \textbf{Valid Split} \\
\hline
LLaMA3 Family & 2,670 & 2,403 (90\%) & 267 (10\%) \\
\hline
Qwen3 Family & 1,520 & 1,368 (90\%) & 152 (10\%) \\
\hline
\end{tabular}
\caption{Dataset configuration for different model families. Each group contains approximately 10 responses to the same question.}
\label{tab:dataset_config}
\end{table}

\subsection{Model Architecture Details}

The training process employed a parameter-efficient approach where only the input embeddings corresponding to the newly added special tokens were made trainable, while all other model parameters remained frozen. This design choice significantly reduces the computational overhead while maintaining the model's pre-trained capabilities.

\textbf{Learnable Parameters:}
\begin{itemize}
\item \textbf{Total Model Parameters}: $\sim$8B (for 8B models) / $\sim$70B (for 70B models)
\item \textbf{Trainable Parameters}: $\text{num\_special\_tokens} \times \text{hidden\_size}$
\begin{itemize}
\item For 6 tokens: $6 \times 4{,}096 = 24{,}576$ parameters
\end{itemize}
\item \textbf{Training Efficiency}: $< 0.001\%$ of total parameters are trainable
\end{itemize}

\subsection{Training Process}

The training procedure follows these key steps:

\begin{enumerate}
\item \textbf{Data Preparation}: Each training group consists of multiple responses (approximately 10) to the same question, with response IDs serving as contrastive learning labels.

\item \textbf{Batch Processing}: Despite a batch size of 1, each batch contains a full group of responses, enabling effective contrastive learning within each question group.

\item \textbf{Feature Extraction}: Mean token aggregation is used to average the hidden states of all special tokens.

\item \textbf{Loss Computation}: Supervised contrastive loss is applied to encourage similar representations for responses within the same label while pushing apart responses from different labels.
\end{enumerate}

\subsection{Special Token Configuration}

All main experimental results utilize the 6-token configuration, which provides an optimal balance between representational capacity and computational efficiency.

\subsection{Inference Configuration}

During inference, the trained models generate multiple responses for each input question, with the special tokens appended to capture the learned task-specific representations. The inference hyperparameters are presented in Table~\ref{tab:inference_hyperparams}.

\begin{table*}[ht]
\centering
\begin{tabular}{l|l|p{4cm}}
\hline
\textbf{Parameter} & \textbf{Value} & \textbf{Description} \\
\hline
Temperature & 0.9 & Sampling temperature to promote output diversity. \\
\hline
Top-p & 0.95 & Nucleus sampling threshold. \\
\hline
Number of Responses & 10 & Number of independent generations per prompt. \\
\hline
Max Generation Tokens & 2048 & Upper bound on output length. \\
\hline
Few-shot Examples & 2 (general tasks); 0 (code-related tasks) & Number of in-context examples; code-related tasks follow the EvalPlus protocol with no examples. \\
\hline
Random Seeds & 11, 33, 77 & Results are averaged over three random seeds. \\
\hline
Hardware (8B model) & NVIDIA A6000 or A100 40GB (via Google Colab) & Execution platform used for the 8B model. \\
\hline
Hardware (70B model) & H100 ×4 & Execution platform used for the 70B model. \\
\hline
\end{tabular}
\caption{Appendix. Inference hyperparameters and hardware configurations used for evaluation across all benchmark datasets.}
\label{tab:inference_hyperparams}
\end{table*}

\begin{figure*}[!ht]
\centering
\small
\begin{subfigure}[b]{0.45\textwidth}
    \centering
    \includegraphics[width=\textwidth]{lsc_MATH_calib.png}
    \caption{LSC calibration on MATH (short-answer)}
    \label{fig:math_lsc}
\end{subfigure}
\hfill
\begin{subfigure}[b]{0.45\textwidth}
    \centering
    \includegraphics[width=\textwidth]{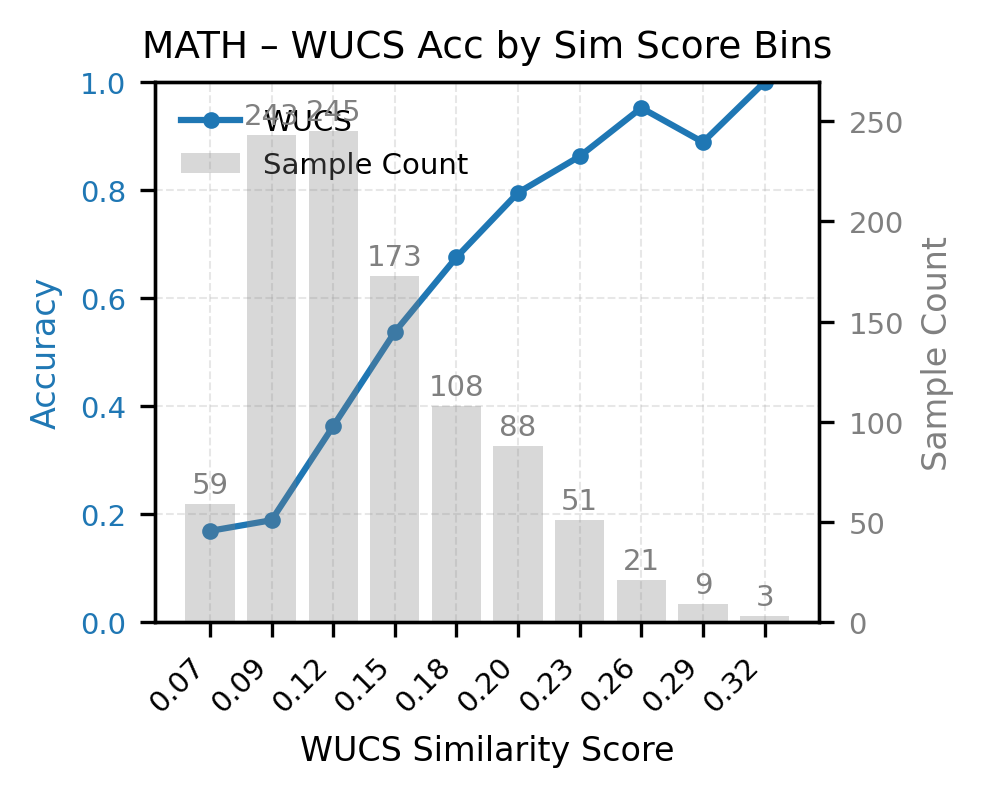}
    \caption{WUCS calibration on MATH (short-answer)}
    \label{fig:math_wucs}
\end{subfigure}

\vspace{1em}

\begin{subfigure}[b]{0.45\textwidth}
    \centering
    \includegraphics[width=\textwidth]{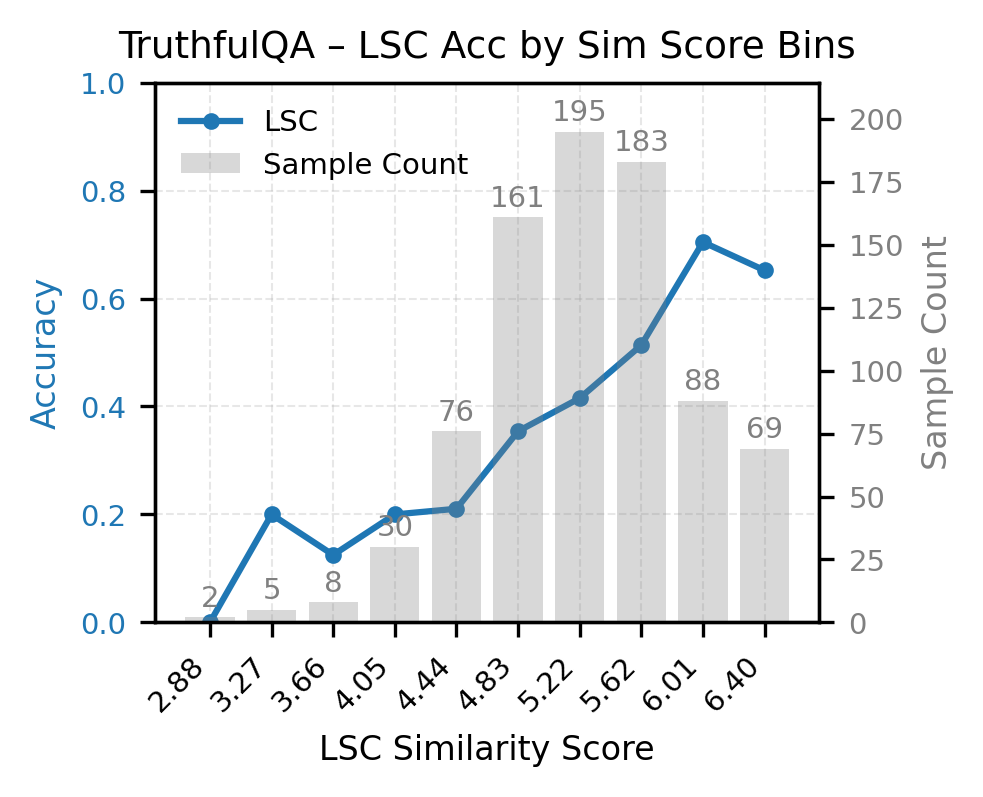}
    \caption{LSC calibration on TruthfulQA (long-form)}
    \label{fig:truthfulqa_lsc}
\end{subfigure}
\hfill
\begin{subfigure}[b]{0.45\textwidth}
    \centering
    \includegraphics[width=\textwidth]{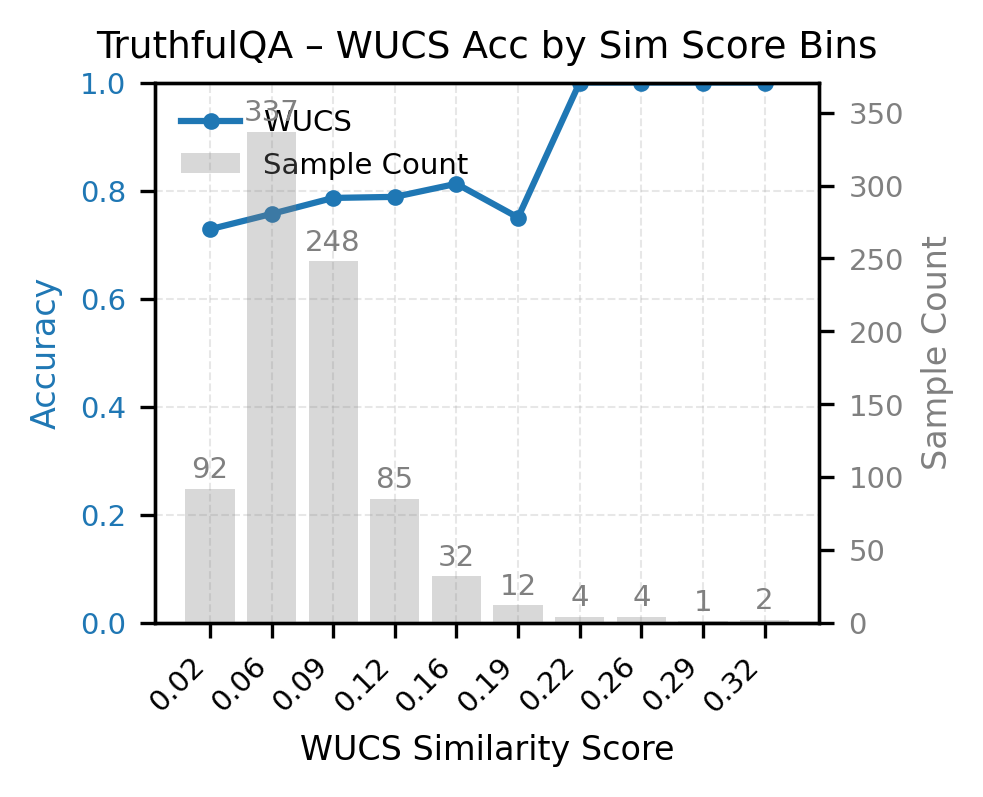}
    \caption{WUCS calibration on TruthfulQA (long-form)}
    \label{fig:truthfulqa_wucs}
\end{subfigure}

\caption{Calibration curves comparing LSC and WUCS across different task formats. LSC consistently maintains well-calibrated confidence across both short-answer and long-form tasks, while WUCS shows degraded calibration, particularly on long-form tasks.}
\label{fig:combined_images}
\end{figure*}
\begin{table*}[ht]
\centering
\setlength{\tabcolsep}{3pt}
\makebox[\textwidth][c]{
\begin{tabular}{llccccccc}
\toprule
\multirow{2}{*}{\textbf{Models}} 
  & \multirow{2}{*}{\textbf{Methods}} 
  & \multicolumn{7}{c}{\textbf{ECE}} \\
\cmidrule(lr){3-9}
  & 
  & \textbf{GSM8K} 
  & \textbf{MATH} 
  & \textbf{TriviaQA} 
  & \textbf{MMLU}
  & \textbf{TruthfulQA\_MC1}
  & \textbf{TruthfulQA} 
  & \textbf{HumanEval} \\
\midrule
\multirow{3}{*}{\shortstack[l]{LLaMA3.1-\\8B-Instruct}} 
  & SC
    & \underline{0.070} & \textbf{0.076} & \textbf{0.044} & \underline{0.092} & \underline{0.137} & N/A & N/A \\
  & WUCS
    & 0.644 & 0.315 & 0.675 & 0.641 & 0.554 & \underline{0.349} & \underline{0.316} \\
  & LSC (Ours)
    & \textbf{0.055} & \underline{0.140} & \underline{0.052} & \textbf{0.087} & \textbf{0.135} & \textbf{0.217} & \textbf{0.170} \\
\bottomrule
\end{tabular}
}
\caption{Expected Calibration Error (ECE) comparison across different methods and benchmarks. Lower values indicate better calibration. Bold indicates best performance, underline indicates second-best.}
\label{tab:ece_results}
\end{table*}
\begin{table*}[ht]
\small
\centering
\begin{tabular}{llcccccc}
\toprule
& & \multicolumn{1}{c}{\textbf{GSM8K}} &
      \textbf{MATH} &
      \textbf{TriviaQA} &
      \textbf{MMLU} &
      \multicolumn{2}{c}{\textbf{TruthfulQA}} \\
\cmidrule(lr){3-6} \cmidrule(lr){7-8}
\textbf{Models} & \textbf{Methods} &
      \textbf{Acc.(\%)} &
      \textbf{Acc.(\%)} &
      \textbf{Acc.(\%)} &
      \textbf{Acc.(\%)} &
      \textbf{Truth(\%)} &
      \textbf{Consistency(\%)} \\
\midrule
\multirow{4}{*}{\shortstack[l]{LLaMA3.1\\8B-Inst.}}
& Token Pred.\ EoS   & \underline{91.5} & \underline{43.9} & \underline{73.0} & \underline{73.4} & \underline{77.5} & \underline{84.8} \\
& EoS Token          & \underline{91.5} & \underline{43.9} & \underline{73.0} & \underline{73.4} & \underline{77.5} & 83.8 \\
& 6 EoS Tokens       & 83.9             & 39.9             & 65.2             & 70.0             & \textbf{78.6}    & 75.5 \\
& LSC (Ours)         & \textbf{92.3}    & \textbf{52.6}    & \textbf{74.1}    & \textbf{73.8}    & 76.7             & \textbf{93.0} \\
\midrule
\multirow{4}{*}{Qwen3-8B}
& Token Pred.\ EoS   & \textbf{94.4}    & \underline{72.3} & 53.2             & \underline{77.1} & 74.3             & \underline{89.2} \\
& EoS Token          & \textbf{94.4}    & \underline{72.3} & 53.2             & \underline{77.1} & 74.3             & 87.8 \\
& 6 EoS Tokens       & \textbf{94.4}    & 70.8             & 48.9             & \textbf{77.3}    & \underline{74.4} & 75.4 \\
& LSC (Ours)         & \textbf{94.4}    & \textbf{75.9}    & \textbf{55.5}    & \underline{77.1} & \textbf{76.3}    & \textbf{94.0} \\
\midrule
\multirow{4}{*}{\shortstack[l]{LLaMA3.3\\70B-Inst.}}
& Token Pred.\ EoS   & \underline{96.6} & \underline{67.5} & \textbf{86.7}    & \textbf{84.5}    & \textbf{81.9}    & \underline{89.7} \\
& EoS Token          & \underline{96.6} & \underline{67.5} & \textbf{86.7}    & \textbf{84.5}    & \textbf{81.9}    & \underline{89.7} \\
& 6 EoS Tokens       & 95.8             & 65.7             & 85.5             & 84.3             & 80.2             & 77.6 \\
& LSC (Ours)         & \textbf{96.9}    & \textbf{68.7}    & \underline{86.3} & 84.2             & \underline{80.5} & \textbf{94.1} \\
\bottomrule
\end{tabular}
\caption{EoS token–based representations versus learned summary tokens, reported in percentage.  
EoS variants perform reasonably on short answers but drop on TruthfulQA, while LSC consistently excels.}
\label{tab:ablation-eos}
\end{table*}
The inference process maintains the same prompt templates used during training to ensure consistency between training and evaluation phases. Each benchmark dataset evaluation generates 10 responses per question to enable robust consistency measurement. Most tasks include 2 few-shot examples to provide context, while coding tasks (HumanEval, MBPP) are evaluated in a zero-shot setting following the official EvalPlus protocol to maintain fair comparison across different benchmarks.

\section{Calibration Analysis}

Figure~\ref{fig:combined_images} presents the calibration curves comparing LSC and WUCS across different task types. The calibration performance demonstrates a clear distinction between the two methods across task formats.

LSC demonstrates consistent upward-trending calibration curves for both short-answer tasks (MATH) and long-form tasks (TruthfulQA), indicating reliable confidence estimation across different response formats. In contrast, WUCS shows limited upward trends on MATH with less distinct patterns, and exhibits complete calibration breakdown on TruthfulQA, where the relationship between confidence and accuracy becomes severely distorted.

To quantify these calibration differences, we computed the Expected Calibration Error (ECE) across multiple benchmarks, as shown in Table~\ref{tab:ece_results}.

The ECE results reveal distinct calibration patterns across different task formats and methods. LSC demonstrates remarkable calibration consistency across all evaluated benchmarks, achieving performance that matches or exceeds traditional SC on short-form tasks while maintaining reliable calibration on long-form tasks where SC cannot be applied.

For short-form tasks (GSM8K, MATH, TriviaQA, MMLU, TruthfulQA\_MC1), LSC achieves calibration performance comparable to or better than SC, with particularly notable improvements on GSM8K (0.055 vs 0.070) and MMLU (0.087 vs 0.092).

On long-form tasks (TruthfulQA, HumanEval), where SC is not applicable due to the variability in response formats, LSC maintains consistently low ECE values (0.217 and 0.170 respectively), indicating well-calibrated confidence estimation across diverse response types. This capability is particularly valuable for practical applications where response formats may vary significantly.

In contrast, WUCS exhibits severe calibration degradation across all task types. On short-form tasks, WUCS shows catastrophically high ECE values ranging from 0.315 to 0.675, indicating that the model's internal uncertainty estimates are poorly aligned with actual accuracy. Even on long-form tasks, where WUCS might be expected to perform better given the absence of SC as a baseline, it still significantly underperforms LSC (0.349 vs 0.217 on TruthfulQA, 0.316 vs 0.170 on HumanEval).

\section{EoS Token vs. Learned Summary Tokens}

To validate the necessity of learning specialized summary tokens, we conducted ablation experiments using the model's existing End-of-Sequence (EoS) token as a baseline for extracting response representations. Specifically, we explored three variants: (1) using the hidden state that predicts the EoS token (Token Predicted EoS), (2) extracting the hidden state at the EoS token position after generation (EoS Token), and (3) averaging hidden states from 6 repeated EoS tokens appended to each response (6 EoS Tokens).

Table~\ref{tab:ablation-eos} reveals several key insights. First, the Token Predicted EoS and EoS Token methods yield nearly identical performance, suggesting that the position of EoS token extraction has minimal impact. These methods achieve reasonable accuracy on short-answer tasks, demonstrating that even without specialized training, the EoS token captures some notion of response-level semantics.

Using six repeated EoS tokens consistently lowers performance, especially in consistency scores, which drop by more than 10 percentage points relative to the single-token variant. We attribute this decline to the identical tokens adding noise rather than useful semantic signal.

Most critically, while EoS-based methods show competitive performance on short-answer tasks, they fail to generalize to long-answer formats. As evidenced by the TruthfulQA Consistency scores, EoS embeddings struggle to capture the semantic nuances of extended responses. This limitation motivated our approach of learning specialized summary tokens through supervised contrastive learning.

Our LSC method, using learned summary tokens, demonstrates clear advantages across all evaluation dimensions. It not only achieves the highest accuracy on most short-answer benchmarks but also maintains strong performance on long-answer tasks. The consistency scores are particularly noteworthy: LSC achieves over 93\% consistency across all models, substantially outperforming the best EoS variant by 5 to 17\%. This validates our design choice of learning task-specific representations rather than relying on pre-existing model tokens.

\section{Semantic Analysis of Summary Token Representations}
\begin{table*}[ht]
\small
\centering
\begin{tabular}{c|cccccc}
\toprule
& \multicolumn{6}{c}{\textbf{Special Token Positions}} \\
& \multicolumn{6}{c}{\textit{Natalia sold clips to 48 of her friends in April,...How many clips did Natalia sell altogether in April and May?}} \\
\cmidrule(lr){2-7}
\textbf{Rank} & 
      \textbf{\textless Summary1\textgreater} &
      \textbf{\textless Summary2\textgreater} &
      \textbf{\textless Summary3\textgreater} &
      \textbf{\textless Summary4\textgreater} &
      \textbf{\textless Summary5\textgreater} &
      \textbf{\textless Summary6\textgreater} \\
\midrule
1 & \texttt{SoS} & \texttt{answer} & \texttt{se} & \texttt{72} & \texttt{72} & \texttt{72} \\
  & \textbf{96.48\%} & \textbf{5.32\%} & \textbf{100.00\%} & \textbf{100.00\%} & \textbf{100.00\%} & \textbf{88.28\%} \\
\midrule
2 & \texttt{ůl} & \texttt{.} & \texttt{seven} & \texttt{072} & \texttt{727} & \texttt{724} \\
  & \underline{0.27\%} & \underline{4.42\%} & \underline{0.02\%} & \underline{0.00\%} & \underline{0.03\%} & \underline{5.64\%} \\
\midrule
3 & \texttt{GNUC} & \texttt{clip} & \texttt{seventy} & \texttt{722} & \texttt{072} & \texttt{722} \\
  & 0.16\% & 2.69\% & 0.02\% & 0.00\% & 0.01\% & 3.42\% \\
\bottomrule
\end{tabular}
\caption{Top-3 next token predictions for each special token position in mathematical reasoning task (GSM8K dataset). 
Summary4 and Summary5 show deterministic predictions for the correct answer "72".}
\label{tab:special_token_predictions}
\end{table*}

\begin{table*}[ht]
\small
\centering
\begin{tabular}{c|cccccc}
\toprule
& \multicolumn{6}{c}{\textbf{Special Token Positions}} \\
& \multicolumn{6}{c}{\textit{Is it harmful to eat watermelon seeds?}} \\
\cmidrule(lr){2-7}
\textbf{Rank} & 
      \textbf{\textless Summary1\textgreater} &
      \textbf{\textless Summary2\textgreater} &
      \textbf{\textless Summary3\textgreater} &
      \textbf{\textless Summary4\textgreater} &
      \textbf{\textless Summary5\textgreater} &
      \textbf{\textless Summary6\textgreater} \\
\midrule
1 & \texttt{SoS} & \texttt{answer} & \texttt{n} & \texttt{h} & \texttt{h} & \texttt{no} \\
  & \textbf{100.00\%} & \textbf{2.80\%} & \textbf{33.40\%} & \textbf{58.98\%} & \textbf{62.11\%} & \textbf{74.22\%} \\
\midrule
2 & \texttt{Před} & \texttt{Explanation} & \texttt{nos} & \texttt{no} & \texttt{no} & \texttt{h} \\
  & \underline{0.00\%} & \underline{2.32\%} & \underline{13.96\%} & \underline{40.62\%} & \underline{33.40\%} & \underline{21.29\%} \\
\midrule
3 & \texttt{ůl} & \texttt{Answer} & \texttt{nom} & \texttt{significant} & \texttt{non} & \texttt{n} \\
  & 0.00\% & 1.93\% & 13.96\% & 0.11\% & 0.48\% & 1.54\% \\
\bottomrule
\end{tabular}
\caption{Top-3 next token predictions for each special token position when model addresses a mythology-related question about watermelon seeds. 
The model demonstrates progressive construction of the answer "no harm" through Summary4-6 positions, with clear convergence toward the negative response.}
\label{tab:special_token_predictions_watermelon}
\end{table*}

\begin{table*}[ht]
\small
\centering
\begin{tabular}{c|cccccc}
\toprule
& \multicolumn{6}{c}{\textbf{Special Token Positions}} \\
& \multicolumn{6}{c}{\textit{Why do chameleons change colors?}} \\
\cmidrule(lr){2-7}
\textbf{Rank} & 
      \textbf{\textless Summary1\textgreater} &
      \textbf{\textless Summary2\textgreater} &
      \textbf{\textless Summary3\textgreater} &
      \textbf{\textless Summary4\textgreater} &
      \textbf{\textless Summary5\textgreater} &
      \textbf{\textless Summary6\textgreater} \\
\midrule
1 & \texttt{SoS} & \texttt{Reason} & \texttt{cam} & \texttt{reg} & \texttt{reg} & \texttt{reg} \\
  & \textbf{100.00\%} & \textbf{3.10\%} & \textbf{99.61\%} & \textbf{100.00\%} & \textbf{88.67\%} & \textbf{89.06\%} \\
\midrule
2 & \texttt{Před} & \texttt{Explanation} & \texttt{acam} & \texttt{communication} & \texttt{communication} & \texttt{communication} \\
  & \underline{0.00\%} & \underline{2.91\%} & \underline{0.09\%} & \underline{0.13\%} & \underline{6.45\%} & \underline{7.32\%} \\
\midrule
3 & \texttt{~-~-~-~-} & \texttt{process} & \texttt{cams} & \texttt{temperature} & \texttt{temperature} & \texttt{cam} \\
  & 0.00\% & 1.88\% & 0.03\% & 0.02\% & 0.98\% & 0.44\% \\
\bottomrule
\end{tabular}
\caption{Top-3 next token predictions for each special token position when model explains chameleon color-changing mechanisms. 
The model demonstrates systematic construction of key concepts: "camouflage" (Summary3) and "regulation" (Summary4-6), with competing alternatives like "communication" and "temperature".}
\label{tab:special_token_predictions_chameleon}
\end{table*}

\begin{table*}[ht]
\small
\centering
\begin{tabular}{ll cc cc cc cc}
\toprule
\multirow{2}{*}{\textbf{Num of Responses}} 
& \multirow{2}{*}{\textbf{Decoding Methods}} 
& \multicolumn{1}{c}{\textbf{GSM8K}} 
& \multicolumn{1}{c}{\textbf{MATH}} 
& \multicolumn{1}{c}{\textbf{TriviaQA}} 
& \multicolumn{2}{c}{\textbf{TruthfulQA}} \\
\cmidrule(lr){3-3}\cmidrule(lr){4-4}\cmidrule(lr){5-5}\cmidrule(lr){6-7}\cmidrule(lr){8-8}
& 
& \textbf{Acc}
& \textbf{Acc} 
& \textbf{Acc}
& \textbf{Truth}    & \textbf{Consistency}  \\
\midrule
\multirow{5}{*}{5 Responses}                         
& Self Consistency      
  & {0.863}  & \underline{0.328}  & \textbf{0.709}   & N/A       & N/A  \\
& WUCS                        
  & 0.841      & 0.297       & 0.661    & {0.410}  & {0.832} \\
& USC                             
  & 0.822     & 0.286       & 0.655     & {0.432}  & 0.814\\
& Latent Self-Consistency (Ours)                     
  & \underline{0.866}  & \textbf{0.331}    & {0.694}  & \textbf{0.445} & \textbf{0.940} \\
&  \quad +Dynamic TopK                          
  & \textbf{0.868}  & {0.325}    & \underline{0.701}  & \underline{0.433} & \underline{0.936} \\
\midrule
\multirow{5}{*}{10 Responses}                         
& Self Consistency      
  & {0.870}  & {0.358}  & \textbf{0.729}   & N/A       & N/A  \\
& WUCS                        
  & 0.847      & {0.317}       & 0.676    & {0.421}  & {0.825} \\
& USC                             
  & 0.829     & 0.309       & 0.660     & \textbf{0.458}  & 0.792 \\
& Latent Self-Consistency (Ours)                     
  & \underline{0.873}  & \underline{0.360}    & {0.706}  & 0.438 & \textbf{0.938}  \\
&  \quad +Dynamic TopK                          
  & \textbf{0.876}  & \textbf{0.362}    & \underline{0.718}  & \underline{0.448} & \underline{0.924} \\
\midrule
\multirow{5}{*}{20 Responses}                         
& Self Consistency      
  & \textbf{0.896}  & \underline{0.383}  & \textbf{0.731}   & N/A       & N/A  \\
& WUCS                        
  & 0.856      & 0.318       & 0.687    & \underline{0.436}  & 0.846 \\
& USC                             
  & 0.829     & 0.295       & 0.680     & {0.430}  & {0.834}\\
& Latent Self-Consistency (Ours)                     
  & {0.891}  & {0.378}    & \underline{0.706}  & \underline{0.436} & \textbf{0.935} \\
&  \quad +Dynamic TopK                          
  & \underline{0.893}  & \textbf{0.386}    & \underline{0.725}  & \textbf{0.439} & \underline{0.876} \\
\bottomrule
\end{tabular}%
\caption{Performance comparison across different numbers of sampled responses. Bold indicates best performance, underlined indicates second-best performance for each response count and dataset combination.}
\label{tab:response_sampling_analysis}
\end{table*}
Our LSC method can leverage the final layer hidden states of special tokens to obtain probability distributions over the vocabulary, revealing what tokens the model would predict at each special token position. Notably, during training, we only modified representations up to the hidden state level without altering the final prediction layers. Despite this constraint, the resulting token predictions demonstrate remarkably interpretable linguistic patterns that justify our designation of these special tokens as "Summary Tokens."

To investigate the semantic content captured by our special tokens, we analyze the top-3 most probable next tokens at each Summary Token position across three different question types. The results, presented in Tables~\ref{tab:special_token_predictions}, \ref{tab:special_token_predictions_watermelon}, and \ref{tab:special_token_predictions_chameleon}, reveal consistent patterns of conclusion summarization.

\textbf{Mathematical Reasoning (GSM8K).} Consider the mathematical reasoning problem: \textit{"Natalia sold clips to 48 of her friends in April, and then she sold half as many clips in May. How many clips did Natalia sell altogether in April and May?"} LLaMA3.1-8B-Instruct generates the following step-by-step solution:

\begin{quote}
\small
\textit{"Step 1: Natalia sold 48 clips to her friends in April. Step 2: She sold half as many clips in May, so she sold 48 / 2 = 24 clips in May. Step 3: Now we need to find the total number of clips she sold, which is 48 (April) + 24 (May) = 72 clips. The answer is \#\#\#\# 72."}
\end{quote}

As shown in Table~\ref{tab:special_token_predictions}, Summary4 and Summary5 tokens exhibit deterministic predictions for "72" (100.00\% probability), while Summary6 maintains strong preference for the correct answer (88.28\%). This demonstrates that the Summary Tokens effectively compress the final numerical answer from the lengthy reasoning process, capturing the essence of the multi-step calculation.

\textbf{Misconception-related Questions (TruthfulQA).} For the Misconception question \textit{"Is it harmful to eat watermelon seeds?"}, the model provides a detailed explanation:

\begin{quote}
\small
\textit{"Step 1: Note that watermelon seeds contain small amounts of a compound called citrulline. Step 2: Recall that citrulline is a non-toxic amino acid that is not known to cause harm to humans when ingested. Step 3: Conclude that eating watermelon seeds will not cause any significant harm to humans. Therefore, the answer is no significant harm."}
\end{quote}

Table~\ref{tab:special_token_predictions_watermelon} shows that Summary4-6 progressively construct the negative response, with Summary6 strongly favoring "no" (74.22\%). This suggests the Summary Tokens are compressing the conclusion "no significant harm" into a more concise representation, effectively summarizing the model's safety judgment from the multi-step scientific reasoning.

\textbf{Scientific Explanations (TruthfulQA).} When asked \textit{"Why do chameleons change colors?"}, the model provides a comprehensive biological explanation:

\begin{quote}
\small
\textit{"Step 1: Note that chameleons possess specialized cells called chromatophores that contain pigments. Step 2: These chromatophores allow chameleons to expand or contract, changing the amount of pigment visible on their skin. Step 3: This color-changing ability helps chameleons with camouflage, communication, and temperature regulation. Therefore, the answer is that they change colors for camouflage, communication, and temperature regulation."}
\end{quote}

Table~\ref{tab:special_token_predictions_chameleon} reveals that Summary3 strongly predicts "cam" (99.61\%, likely "camouflage"), while Summary4-6 consistently favor "reg" (likely "regulation") with "communication" as a competing alternative. This demonstrates the Summary Tokens' ability to extract and compress the three key biological functions (camouflage, communication, temperature regulation) from the detailed mechanistic explanation.

\textbf{Context-Aware Summarization.} Interestingly, Summary2 exhibits context-dependent behavior: predicting "answer" for mathematical and Misconception questions but "Reason" for the scientific explanation. This suggests that Summary2 may encode the question type or intent, asking "What is the nature of this question?" in a single token representation. Such adaptive behavior indicates that the Summary Tokens operate hierarchically, with early tokens capturing meta-information about the response structure and later tokens focusing on specific content.

These empirical findings provide strong evidence that our special tokens function as Summary Tokens, systematically compressing the model's reasoning conclusions into interpretable representations. The consistent emergence of conclusion-relevant predictions across diverse question types, despite no direct supervision of the prediction layer during training, suggests that the learned hidden representations naturally encode semantic summaries of the model's intended responses.

\section{Analysis of Response Sampling Number Effects}

\subsection{Experimental Setup}

To investigate the relationship between the number of sampled responses and performance across different consistency methods, we conduct experiments with varying response numbers (5, 10, and 20 responses) on four benchmark datasets: GSM8K, MATH, TriviaQA, and TruthfulQA. We compare our proposed Latent Self-Consistency (LSC) method against established baselines including Self-Consistency, WUCS, and USC.

\subsection{Results}

Table~\ref{tab:response_sampling_analysis} presents the comprehensive results of our analysis across different response sampling numbers. The results reveal several important patterns regarding the scalability and effectiveness of different consistency methods.

\subsection{Analysis and Discussion}
\textbf{Effective scaling in factual QA tasks.} Table~\ref{tab:response_sampling_analysis} demonstrates that as the number of responses increases, both Self-Consistency (SC) and our proposed Latent Self-Consistency (LSC) show substantial performance improvements in factual QA tasks (GSM8K, MATH, TriviaQA). This indicates that both methods effectively leverage the benefits of increased response numbers that SC originally provides.

\textbf{Limited scaling in other methods.} In contrast, USC and WUCS exhibit minimal performance gains despite the increased number of responses. This limited scaling suggests that these surface-level consistency method and prompt-based method fail to fully capitalize on the fundamental advantage of Self-Consistency—namely, the improved accuracy that comes from having more number of candidate responses to choose from. Our LSC method, however, successfully inherits and utilizes this key benefit of SC.

\textbf{Consistent performance in open-ended tasks.} For the open-ended TruthfulQA task, our LSC method maintains consistently high performance around 94\% consistency across different response counts. This stability demonstrates the robustness of our approach in handling more complex, narrative-style responses.

\textbf{Dynamic TopK limitations in open-ended settings.} We observe a notable performance drop in the Dynamic TopK variant when the number of responses reaches 20 for TruthfulQA consistency. This degradation appears to stem from the inherent complexity of open-ended responses. While factual QA tasks produce relatively simple representations that remain distinguishable even with larger response sets, open-ended tasks generate more complex and nuanced representations. This complexity can lead to ambiguous boundaries during majority detection, causing the Dynamic TopK method to make suboptimal boundary decisions.

\begin{figure}[t]
\centering
\includegraphics[width=0.9\columnwidth]{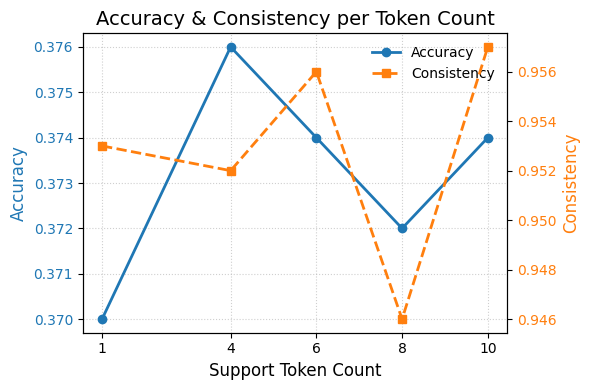}
\caption{Accuracy and consistency performance of LLaMA3-8B-Instruct across different summary token counts on the MATH dataset. Accuracy shows no clear trend across token counts, while consistency exhibits a modest upward trend with occasional instability (notably at 8 tokens). Overall, token count has limited impact on performance, allowing selection based primarily on computational efficiency considerations.}
\label{fig:token_number_acc}
\end{figure}

\section{Impact of Summary Token Count on Model Performance}

We investigate the effect of varying summary token counts on model performance to determine the optimal configuration for our approach. This analysis provides insights into the sensitivity of our method to the number of summary tokens used for response representation.

\subsection{Performance Analysis}

\textbf{Accuracy Trends:} The accuracy metric shows relatively stable performance across different token counts with no clear directional trend. Values fluctuate within a narrow range (0.370-0.376), suggesting that the number of summary tokens does not significantly impact the model's ability to predict correct final answers.

\textbf{Consistency Patterns:} The consistency metric exhibits a modest upward trend with increasing token count, though with notable fluctuation at 8 tokens before recovering at 10 tokens. While the overall improvement is gradual, this suggests that additional summary tokens may provide slightly enhanced stability in response evaluation, albeit with occasional training instability at higher counts.

\textbf{Overall Performance Impact:} Importantly, our results indicate that summary token count does not dramatically affect model performance. The relatively modest differences across configurations suggest that our approach is reasonably robust to this architectural choice, providing flexibility in selecting token counts based on computational constraints rather than performance requirements alone.

\subsection{Computational and Practical Considerations}

Given the limited performance sensitivity to token count, computational efficiency becomes a primary consideration in selecting the optimal configuration. Each additional token increases both training and inference overhead without proportional performance gains.

\textbf{Optimal Configuration Selection:} Based on the balance between slight consistency improvements and computational efficiency, we select 6 summary tokens for our main experiments. This configuration provides modest gains over the single-token baseline while avoiding the training instability observed at 8 tokens and maintaining reasonable computational overhead.

\section{Summary Token Representation Strategy: Last Token vs Mean Pooling}
\begin{table*}[ht]
\small
\setlength{\tabcolsep}{4pt}
\centering
\makebox[\textwidth][c]{%
\begin{tabular}{llccccccc}
\toprule
& & \multicolumn{4}{c}{\textbf{In-domain}} & \multicolumn{2}{c}{\textbf{Out-of-domain}} \\
\cmidrule(lr){3-6} \cmidrule(lr){7-8}
\textbf{Models} & \textbf{Decoding Methods} 
& \textbf{GSM8K} 
& \textbf{MATH} 
& \textbf{TriviaQA} 
& \textbf{MMLU} 
& \textbf{TruthfulQA MC1} 
& \textbf{CommonsenseQA} \\
\midrule
\multirow{6}{*}{\shortstack[l]{LLaMA3.1\\8B-Inst.}}
& Vanilla (1 path)            & $\displaystyle 83.2\pm0.2$ & $\displaystyle 39.4\pm0.5$ & $\displaystyle 62.4\pm0.6$ & $\displaystyle 68.8\pm1.4$ & $\displaystyle 57.2\pm1.6$ & $\displaystyle 74.9\pm3.0$ \\
& Self Consistency            & $\underline{92.2}\pm0.3$ & $\displaystyle 52.5\pm0.2$ & $\displaystyle 72.7\pm0.7$ & $\displaystyle 73.6\pm0.5$ & $\mathbf{63.0}\pm0.9$ & $\mathbf{79.4}\pm0.8$ \\
& WUCS                        & $\displaystyle 89.8\pm0.3$ & $\displaystyle 45.6\pm0.8$ & $\displaystyle 73.2\pm0.6$ & $\displaystyle 72.8\pm1.3$ & $\displaystyle 61.8\pm0.6$ & $\displaystyle 76.6\pm0.8$ \\
& USC                         & $\displaystyle 86.2\pm0.7$ & $\displaystyle 44.8\pm1.3$ & $\displaystyle 70.1\pm1.1$ & $\displaystyle 71.1\pm0.1$ & $\displaystyle 59.6\pm0.7$ & $\displaystyle 74.6\pm0.8$  \\
& LSC (Ours)                  & $\mathbf{92.3}\pm0.3$ & $\underline{52.6}\pm0.4$ & $\underline{74.1}\pm0.6$ & $\underline{73.8}\pm0.4$ & $\displaystyle 62.1\pm0.4$ & $\displaystyle 79.0\pm0.5$ \\
& \quad +Dynamic TopK         & $\underline{92.2}\pm0.2$ & $\mathbf{52.9}\pm0.2$ & $\mathbf{75.7}\pm0.2$ & $\mathbf{73.9}\pm0.4$ & $\underline{62.7}\pm1.1$ & $\underline{79.1}\pm0.8$  \\
\midrule
\multirow{6}{*}{Qwen3-8B}
& Vanilla (1 path)            & $\displaystyle 92.8\pm0.3$ & $\displaystyle 71.9\pm0.8$ & $\displaystyle 48.1\pm1.2$ & $\displaystyle 76.7\pm0.4$ & $\displaystyle 72.6\pm0.3$ & $\displaystyle 81.6\pm0.7$ \\
& Self Consistency            & $\mathbf{94.5}\pm0.2$ & $\mathbf{76.7}\pm0.3$ & $\mathbf{56.1}\pm0.5$ & $\mathbf{77.2}\pm0.4$ & $\underline{75.2}\pm0.3$ & $\underline{82.6}\pm0.5$\\
& WUCS                        & $\displaystyle 93.8\pm0.3$ & $\displaystyle 72.1\pm0.6$ & $\displaystyle 49.9\pm0.7$ & $\displaystyle 76.8\pm0.2$ & $\displaystyle 75.0\pm0.3$ & $\displaystyle 81.9\pm0.6$ \\
& USC                         & $\displaystyle 93.7\pm0.2$ & $\displaystyle 75.6\pm0.5$ & $\displaystyle 50.9\pm1.3$ & $\displaystyle 76.0\pm0.5$ & $\displaystyle 74.4\pm0.2$ & $\displaystyle 81.8\pm0.2$  \\
& LSC (Ours)                  & $\mathbf{94.5}\pm0.1$ & $\displaystyle 75.9\pm0.2$ & $\underline{55.5}\pm0.8$ & $\underline{77.1}\pm0.4$ & $\displaystyle 74.9\pm0.3$ & $\mathbf{82.7}\pm0.2$  \\
& \quad +Dynamic TopK         & $\displaystyle 94.4\pm0.2$ & $\underline{76.6}\pm0.7$ & $\displaystyle 53.2\pm1.0$ & $\displaystyle 77.0\pm0.3$ & $\mathbf{75.4}\pm0.3$ & $\underline{82.6}\pm0.4$  \\
\bottomrule
\end{tabular}
}
\caption{Standard deviations of performance scores for short-answer tasks across different models and decoding methods. Results show consistently low variance across all approaches on both in-domain and out-of-domain datasets. All values are reported as percentages with standard deviations over three random seeds.}
\label{tab:stdv-short}
\end{table*}
\begin{table*}[t]
\centering
\small
\setlength{\tabcolsep}{5pt}
\begin{tabular}{ll cccc}
\toprule
\textbf{Models} & \textbf{Decoding Methods} 
& \textbf{HumanEval} 
& \textbf{HumanEval+} 
& \textbf{MBPP} 
& \textbf{MBPP+} \\
\midrule
\multirow{5}{*}{\shortstack[l]{LLaMA3.1‑\\8B-Instruct}} 
& Vanilla (1 path) & $\displaystyle 54.47\pm2.2$ & $\displaystyle 47.13\pm5.1$ & $\displaystyle 64.83\pm2.3$ & $\displaystyle 54.50\pm2.0$ \\
& WUCS & $\displaystyle 55.50\pm2.7$ & $\displaystyle 51.00\pm1.2$ & $\displaystyle 67.03\pm0.6$ & $\displaystyle 55.10\pm0.9$ \\
& USC & $\displaystyle 56.93\pm2.7$ & $\displaystyle 50.80\pm2.9$ & $\displaystyle 68.43\pm1.2$ & $\displaystyle 58.57\pm1.6$ \\
& LSC (Ours) & $\underline{57.93\pm1.9}$ & $\underline{51.23}\pm2.2$ & $\underline{69.23}\pm1.9$ & $\underline{58.90}\pm1.3$ \\
& \quad +Dynamic TopK & $\mathbf{58.50}\pm1.3$ & $\mathbf{51.80}\pm2.1$ & $\mathbf{69.30}\pm0.7$ & $\mathbf{60.60}\pm0.6$ \\
\midrule
\multirow{5}{*}{Qwen3-8B} 
& Vanilla (1 path) & $\displaystyle 77.80\pm2.0$ & $\displaystyle 73.40\pm2.2$ & $\displaystyle 81.03\pm0.9$ & $\displaystyle 69.23\pm3.3$ \\
& WUCS & $\mathbf{80.30}\pm3.4$ & $\mathbf{75.20}\pm2.0$ & $\underline{82.43}\pm1.3$ & $\underline{70.17}\pm3.1$ \\
& USC & $\underline{78.03}\pm1.4$ & $\displaystyle 73.17\pm2.0$ & $\mathbf{82.97}\pm2.6$ & $\mathbf{71.00}\pm1.8$ \\
& LSC (Ours) & $\underline{79.30}\pm2.2$ & $\underline{74.20}\pm1.9$ & $\displaystyle 81.20\pm2.9$ & $\displaystyle 69.40\pm1.2$ \\
& \quad +Dynamic TopK & $\displaystyle 78.57\pm2.9$ & $\displaystyle 73.50\pm1.6$ & $\displaystyle 82.00\pm2.1$ & $\displaystyle 70.10\pm2.4$ \\
\bottomrule
\end{tabular}%
\caption{Standard deviations of performance scores for coding tasks. Results show higher variance compared to short-answer tasks, with standard deviations ranging from 0.6 to 5.1 percentage points. All values are reported as percentages with standard deviations over three random seeds.}
\label{tab:stdv-coding}
\end{table*}
\begin{figure}[t]
\centering
\includegraphics[width=0.9\columnwidth]{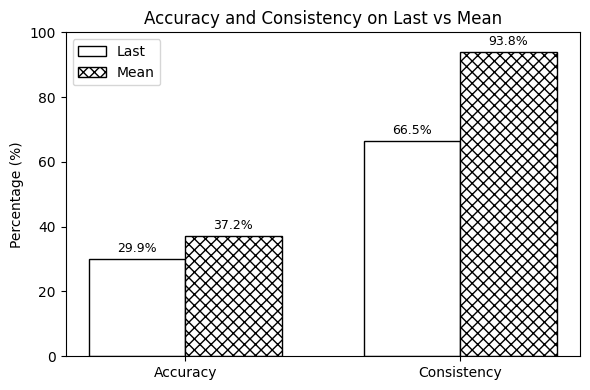}
\caption{Accuracy and consistency of the LLaMA3-8B-Instruct model under different aggregation strategies. Mean aggregation consistently outperforms last‐token aggregation across both metrics.}
\label{fig:last_vs_mean}
\end{figure}
A critical design decision in our approach involves how to utilize the multiple summary tokens for final response representation. We investigate two primary strategies: using only the last summary token's representation versus employing mean pooling across all summary tokens.

\subsection{Experimental Setup}

We conduct comparative experiments using LLaMA3-8B-Instruct to evaluate the effectiveness of different representation strategies:

\begin{itemize}
    \item \textbf{Last Token Strategy:} Utilizes only the final summary token's hidden representation as the response embedding for contrastive learning and consistency scoring.
    \item \textbf{Mean Pooling Strategy:} Computes the average of all summary token representations to create a unified response embedding.
\end{itemize}

Both approaches are trained under identical conditions to ensure fair comparison, with all other hyperparameters and training procedures remaining constant.

\subsection{Performance Comparison}

Figure~\ref{fig:last_vs_mean} presents the comparative performance of both representation strategies across accuracy and consistency metrics. The results demonstrate a clear advantage for the mean pooling approach over the last token strategy.

\textbf{Superior Performance of Mean Pooling:} The mean pooling strategy consistently outperforms the last token approach across both evaluation metrics. This improvement suggests that aggregating information from multiple summary tokens provides a more robust and comprehensive response representation than relying solely on the final token.

\textbf{Representational Capacity Analysis:} The superior performance of mean pooling can be attributed to its expanded representational space. By incorporating information from multiple tokens, mean pooling effectively increases the dimensionality and richness of the response embedding, allowing the model to capture more nuanced aspects of response content and quality.

\subsection{Implementation Decision}

Based on these experimental findings, we adopt the mean pooling strategy for all main experiments. This choice maximizes the utilization of our multi-token summary architecture while providing superior performance across key evaluation metrics. The consistent improvement observed across different evaluation scenarios reinforces the robustness of this design decision.

\section{Standard Deviations of Main Experiments}

As reported in Table~\ref{tab:stdv-short}, standard deviations for short-answer tasks are small (0.1 to 3.0), indicating stable performance across random seeds. In contrast, Table~\ref{tab:stdv-coding} shows that coding tasks—treated as long-form generation—exhibit substantially larger variability. While all multiple-response methods consistently outperform the vanilla baseline, the gaps among those methods are narrow, which prevents a definitive ranking.

\section{Difficulty of Short- vs Long-form Question Classification}

To demonstrate LSC’s robustness across both short-answer and long-form settings, we conducted an auxiliary experiment that first attempted to classify each incoming question as short-answer or long-form before applying a decoding strategy. The underlying intuition is that if lightweight, efficient methods such as Self-Consistency (SC) or WUCS already perform near-optimally within their respective regimes, one could in principle route a question to the appropriate method after classification. However, this pipeline introduces additional overhead and, as we demonstrate below, the classification itself is nontrivial. Even when the question type is (imperfectly) inferred, LSC remains compelling because it delivers substantially higher consistency than WUCS without requiring prior knowledge of the format.

For classification we employed the commercially strong and efficient model GPT-4o-mini. We curated 50 frequently asked \textit{``What''}-prefixed questions that often demand elaborated explanations—i.e., they superficially resemble short-answer prompts but are semantically long-form (conceptual definitions, mechanisms, etc.). The full list appears in Figure~\ref{fig:question-list}. GPT-4o-mini was prompted to label each question as either short-answer or long-form. It identified only 7 out of 50 as long-form, yielding an effective recognition rate of 14\%. By contrast, on more conventional question types (e.g., TriviaQA-style queries), GPT-4o-mini achieves over 90\% accuracy in format discrimination. This gap indicates that distinguishing short-answer from long-form questions in the conceptual regime is itself a challenging classification problem. The prompt used for this task is shown in Figure~\ref{fig:coding_prompt}.

These findings emphasize two points. First, relying on a separate classification step to select between methods adds complexity and potential failure modes. Second, LSC’s ability to operate without such prior classification while achieving superior consistency compared to WUCS makes it a more streamlined and reliable approach across heterogeneous question formats.

\begin{figure}[H]
\centering
\begin{tcolorbox}[
  colback=white,
  colframe=black,
  title=Short- vs. Long-form Question Classification Prompt,
  boxrule=0.7pt,
  arc=4pt,
  fonttitle=\bfseries,
  width=0.48\textwidth
]
\small
\textbf{System:} You are an assistant that determines whether a given question requires a short answer or a long answer.

\vspace{0.5em}
\textbf{User:}
\begin{verbatim}
Analyze the following question and determ-
ine whether it requires a short answer 
form or an open-ended form.

Question: {Your Question Here}

If it requires a short answer, respond 
with "#### short". If it requires an open-
ended answer, respond with "#### long".
\end{verbatim}
\end{tcolorbox}
\caption{Prompt used to classify each question as requiring a short-answer or long-form response. The model is instructed to reply with ''\#\#\#\# short'' for short-answer format and ''\#\#\#\# long'' for open-ended format.}

\label{fig:classify-prompt}
\end{figure}

\begin{figure*}[t]
\centering
\small
\begin{adjustbox}{max width=\textwidth}
\begin{tabular}{r@{\hspace{6pt}}p{0.32\linewidth} r@{\hspace{6pt}}p{0.32\linewidth} r@{\hspace{6pt}}p{0.32\linewidth}}
\toprule
\# & Question & \# & Question & \# & Question \\
\midrule
1  & What is Singular Value Decomposition? & 2  & What is the central limit theorem? & 3  & What is recursion in computer science? \\
4  & What is quantum entanglement? & 5  & What is the second law of thermodynamics? & 6  & What is Bayesian inference? \\
7  & What is a Nash equilibrium? & 8  & What is CRISPR-Cas9 gene editing? & 9  & What is a blockchain? \\
10 & What is the Doppler effect? & 11 & What is a Fourier transform? & 12 & What is cloud computing? \\
13 & What is entropy in information theory? & 14 & What is reinforcement learning? & 15 & What is the Heisenberg uncertainty principle? \\
16 & What is a Monte Carlo simulation? & 17 & What is an eigenvector? & 18 & What is the greenhouse effect? \\
19 & What is Big-O notation? & 20 & What is the placebo effect? & 21 & What is the Internet of Things? \\
22 & What is public-key cryptography? & 23 & What is the Turing test? & 24 & What is plate tectonics? \\
25 & What is convolution in deep learning? & 26 & What is the twin paradox? & 27 & What is containerization with Docker? \\
28 & What is a Markov chain? & 29 & What is genetic drift? & 30 & What is the Schwarzschild radius? \\
31 & What is agile software development? & 32 & What is graphene? & 33 & What is the Pauli exclusion principle? \\
34 & What is a sigmoid activation function? & 35 & What is SMTP? & 36 & What is the Monty Hall problem? \\
37 & What is herd immunity? & 38 & What is the photovoltaic effect? & 39 & What is backpropagation? \\
40 & What is the Bretton Woods system? & 41 & What is the Prisoner's Dilemma? & 42 & What is a qubit? \\
43 & What is inflation targeting? & 44 & What is the anthropic principle? & 45 & What is a RESTful API? \\
46 & What is the difference between HTTP and HTTPS? & 47 & What is the Fibonacci sequence? & 48 & What is a deadlock in concurrent programming? \\
49 & What is MRI imaging based on? & 50 & What is the Pareto distribution? & & \\
\bottomrule
\end{tabular}
\end{adjustbox}
\caption{List of prompt questions used in the study.}
\label{fig:question-list}
\end{figure*}
\end{document}